\documentclass[]{journal}

\newcommand{\bump}[0]{\operatorname{bump}}
\newcommand{\fro}{\text{HS}}
\newcommand{\edit}[1]{#1}

\begin{document}

    \begin{frontmatter}

        \title{QuadConv: Quadrature-Based Convolutions with Applications to Non-Uniform PDE Data Compression}

        \author[1]{Kevin Doherty\corref{cor}\fnref{*}}
            \ead{kevin.doherty@colorado.edu}
        \author[2]{Cooper Simpson\fnref{*}}
            \ead{cooper.simpson@colorado.edu}
        \author[1]{Stephen Becker}
            \ead{stephen.becker@colorado.edu}
        \author[2]{Alireza Doostan}
            \ead{doostan@colorado.edu}

        \affiliation[1]{organization={Applied Mathematics},
            addressline={University of Colorado Boulder},
            postcode={80309},
            city={Boulder},
            country={USA}}

        \affiliation[2]{organization={Smead Aerospace Engineering Sciences},
            addressline={University of Colorado Boulder},
            postcode={80309},
            city={Boulder},
            country={USA}}

        \cortext[cor]{Corresponding author}
        \fntext[*]{These authors contributed equally}

        \begin{abstract}
            We present a new convolution layer for deep learning architectures which we call QuadConv --- an approximation to continuous convolution via quadrature. Our operator is developed explicitly for use on non-uniform, mesh-based data, and accomplishes this by learning a continuous kernel that can be sampled at arbitrary locations. Moreover, the construction of our operator admits an efficient implementation which we detail and construct. \edit{As an experimental validation of our operator, we consider the task of compressing partial differential equation (PDE) simulation data from fixed meshes}. We show that QuadConv can match the performance of standard discrete convolutions on uniform grid data by comparing a QuadConv autoencoder (QCAE) to a standard convolutional autoencoder (CAE). Further, we show that the QCAE can maintain this accuracy even on non-uniform data. \edit{In both cases, QuadConv also outperforms alternative unstructured convolution methods such as graph convolution.}
        \end{abstract}


        \begin{keyword}
            Quadrature \sep Convolution \sep Deep-Learning \sep Compression
        \end{keyword}

        \date{\today}

    \end{frontmatter}
	

    \section{Introduction}\label{sec:intro}
    
\begin{figure}[t]
            \centering
            \begin{subfigure}[b]{0.31\textwidth}
                \centering
                \includegraphics[width=0.7\textwidth]{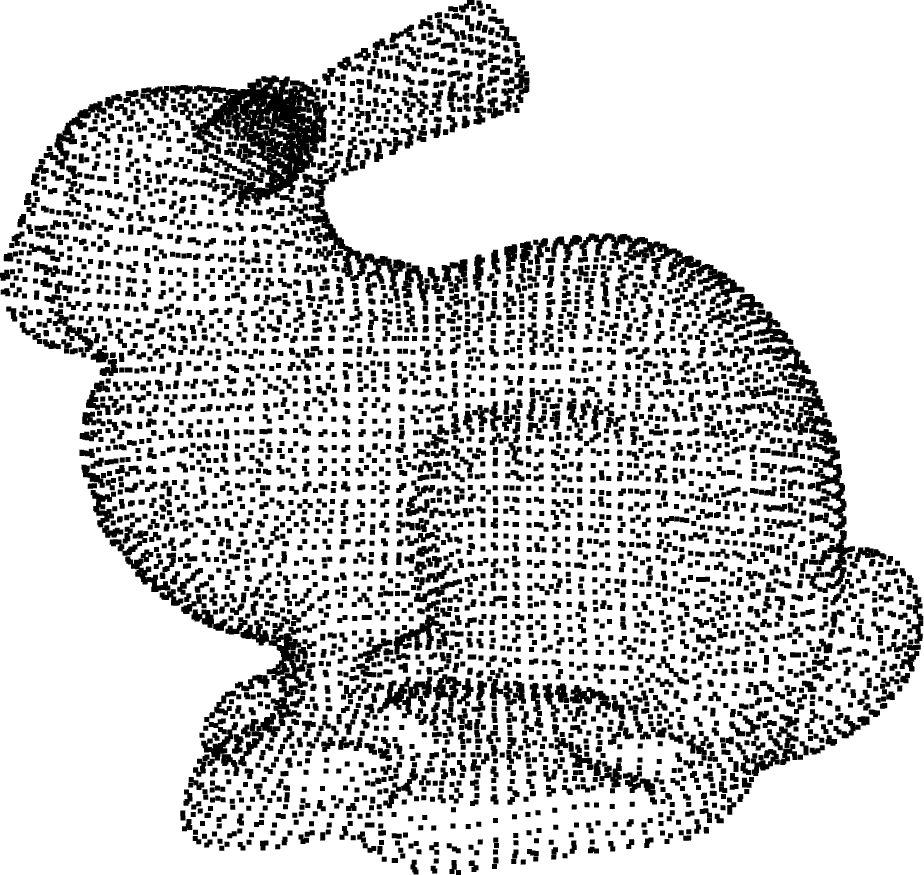}
                \caption{\small 3D point cloud \cite{stanford-bunny}.} 
                \label{fig:pointcloud-ex}
            \end{subfigure}
            \hfill
            \begin{subfigure}[b]{0.31\textwidth}
                \centering
                \includegraphics[width=0.7\textwidth]{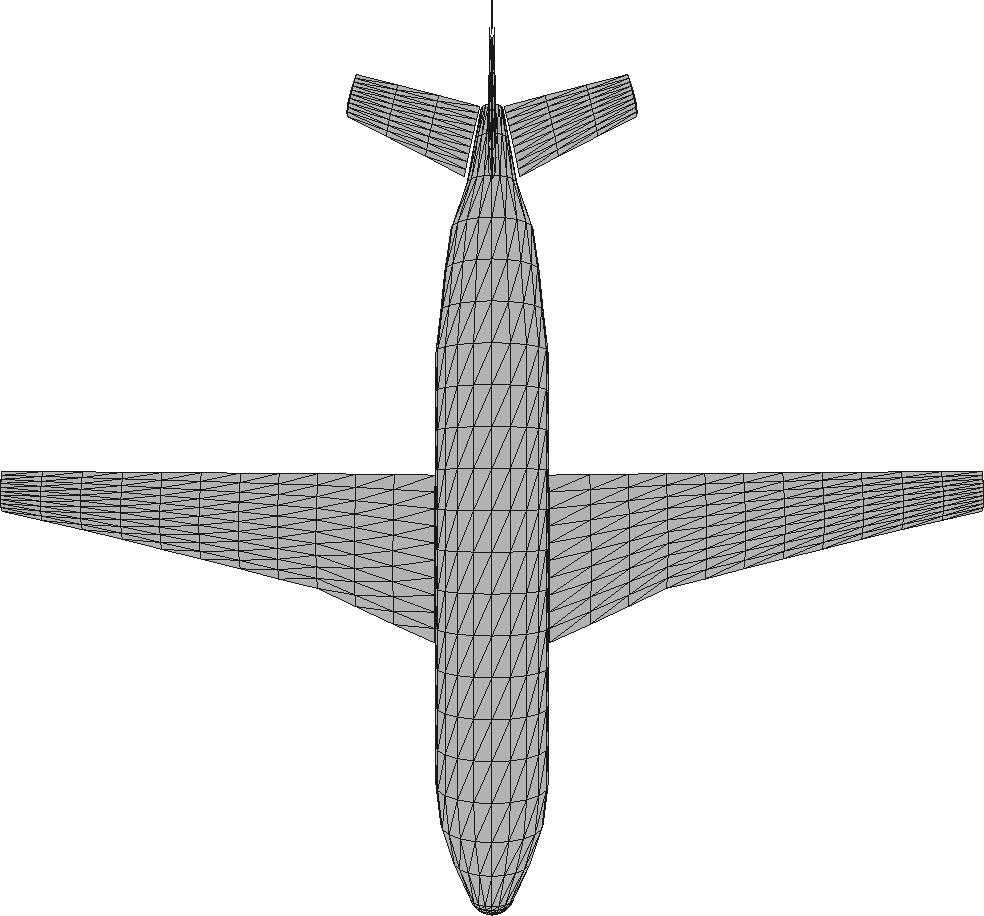}
                \caption{\small Surface mesh \cite{airplane}.}
                \label{fig:mesh-ex-1}
            \end{subfigure}
            \hfill
            \begin{subfigure}[b]{0.31\textwidth}
                \centering
                \includegraphics[width=\textwidth]{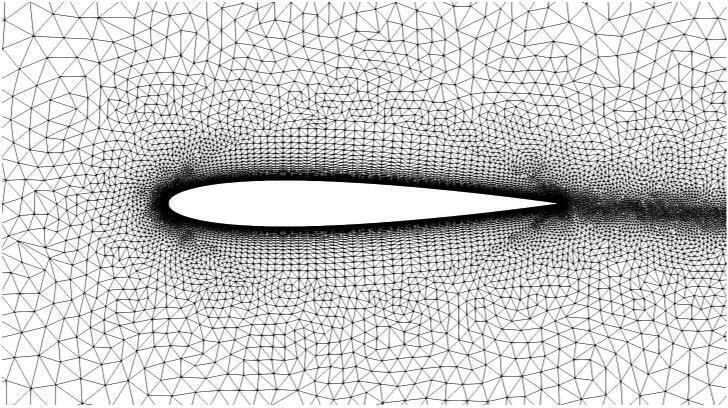}
                \caption{\small NACA 0012 mesh \cite{skinner2019reduced}.} 
                \label{fig:mesh-ex-2}
            \end{subfigure}
            \caption{\small Examples of non-uniform data.}
            \label{fig:non-uniform-ex}
        \end{figure}

        Discrete convolutions are one of the canonical operations used in numerous deep learning applications such as image classification, object detection, semantic segmentation, etc. They have been proven to be effective in extracting important features from data, and they possess a number of desirable properties such as translation equivariance. In particular, when used with compactly supported kernels of fixed size, this facilitates the extraction of local features from the data and provides a significant boost to computational efficiency compared to a (square) fully-connected layer, reducing time complexity from quadratic in the input size to linear and memory from quadratic to a constant \cite{Goodfellow-et-al-2016}. The effects of this are significant: each training epoch is faster, less overall training is required, and often results in better generalization with fewer parameters. However, these standard discrete convolutions rely on the assumption that the data is defined on a uniform grid. This limitation is unfortunate, as there are a host of settings where convolutions may be effective, but the relevant data is non-uniform; \cref{fig:non-uniform-ex} shows a few representative examples.

        In this paper we introduce a quadrature-based discrete convolution operator suitable for arbitrary meshes, which we call QuadConv. Our construction is based on the continuous definition of convolution, which, for two functions \(f,g:\reals^D\to\reals\), is given by
        \begin{equation}\label{eq:ctsconv}
            (f \ast g)(\bm{y})=\int_{\reals^D} f(\bm{x})\cdot g(\bm{y}-\bm{x}) \ d\bm{x}.
        \end{equation}
        A variety of conditions on \(f\) and \(g\) can guarantee this integral is well defined, e.g., Young's Inequality, or, of particular relevance here, if \(f\) and \(g\) are compactly supported. Note that we use the term quadrature to refer to a weighted sum approximation of an integral. Some authors make a distinction between one dimensional quadrature and higher dimensional cubature which we will not employ. In general, our method is applicable to data on a mesh, which is a set of nodes with arbitrary non-intersecting connections. Within this setting, we will consider two particular sub-types of data: uniform grids and non-uniform meshes. The latter explicitly requires the nodes to be non-uniformly distributed in space. Our proposed method is mathematically quite simple, but, as we will see, is non-trivial to implement efficiently.

        We will denote vectors using lowercase bold font (e.g., \(\bm{x}\)), and matrices or operators using uppercase bold font (e.g., \(\bm{X}\)). We will often refer to \(g\) as the kernel and \(f\) as the data. This distinction is important in this context, as the data, \(f\), is given (and hence known only on the mesh nodes) and the kernel, \(g\), is a learned map. Where appropriate, continuous functions will be referred to with an argument from their domain in parenthesis, \(f(\cdot)\) or \(f(\bm{x})\), and its discrete counterpart will be referred to as \(f(\bm{x}_i)\), or, if the arguments are clear from context, with an index \(f_i\).
        
        To summarize our contributions, we propose a novel convolution operator for deep learning applications that is suitable for data on a non-uniform mesh. In addition to this, we discuss the practical implementation of our operator, showing that it is a computationally feasible approach. Lastly, we present the application of our work to autoencoder based data compression. We show that it matches the effectiveness of standard convolutions on uniform grid data, and performs equally well on non-uniform data. 

        In the remainder of this section we focus on motivating our approach and discussing the related literature. \Cref{sec:methods} will then introduce our mathematical formulation, with \cref{sec:computation} describing the practical implementation. In \cref{sec:experiments} we  apply our method to a number of datasets on uniform grids and non-uniform meshes. \Cref{sec:discussion} summarizes our work and presents possible directions for future research.
        
        
        \subsection{Motivation}\label{sec:motivation}
        
\begin{figure}[H]
                \centering
                \begin{subfigure}[b]{\textwidth}
                    \includegraphics[width=\textwidth]{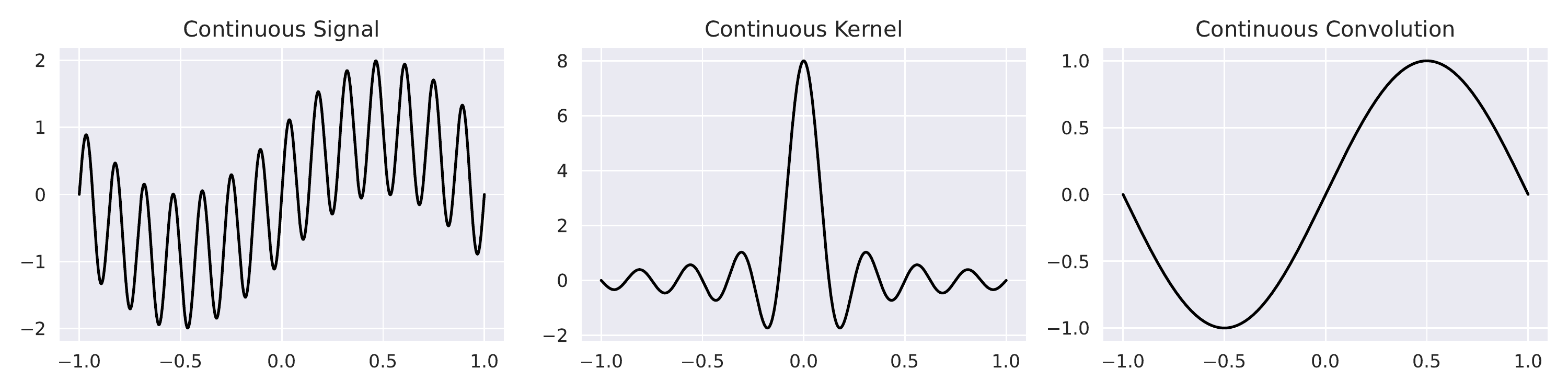}
                    \caption{\small Analytic convolution of continuous kernel and signal.}
                    \label{fig:conv-ex-continuous}
                \end{subfigure}
                \vfill
                \begin{subfigure}[b]{\textwidth}
                    \centering
                    \includegraphics[width=\textwidth]{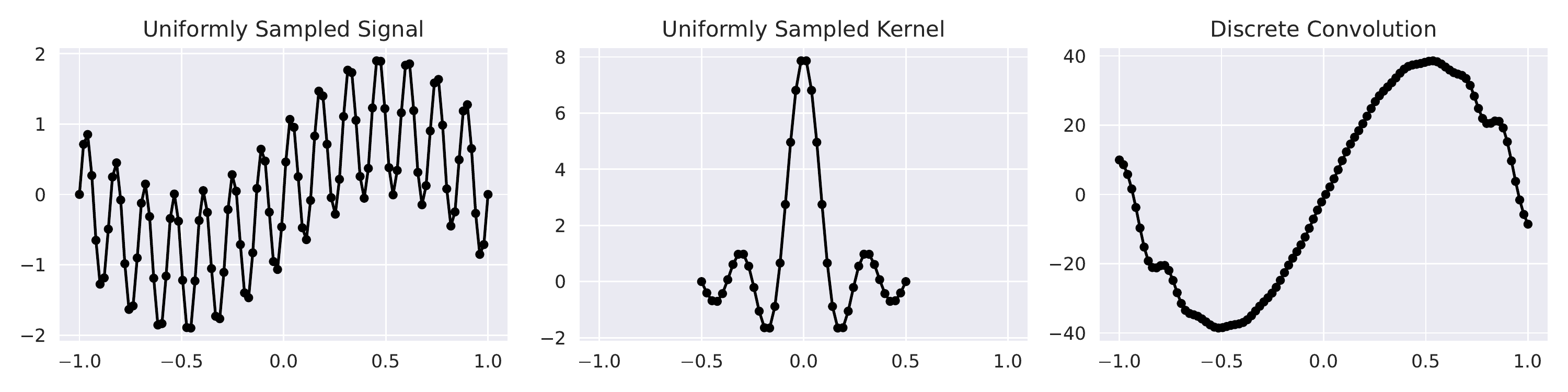}
                    \caption{\small Discrete convolution of uniformly sampled signal and kernel.} 
                    \label{fig:conv-ex-uniform-sampling}
                \end{subfigure}
                \caption{\small Comparison of continuous and discrete convolution.}
                \label{fig:conv-motivation-1}
            \end{figure}

            The standard convolution operator used in deep learning is a particular discretization of \cref{eq:ctsconv} operating on \(D\)-dimensional tensors. For example, for a single output and input channel, the one dimensional form is given as follows:
            \begin{equation}\label{eq:dscconv}
                (f\ast g)_j=\sum_i f_i\cdot g_{j-i}
            \end{equation}
            along with appropriate boundary conditions.
            This discretization is mathematically justified as long as the points are equally spaced, and is drawn from the definition of convolution for functions defined on \(\mathbb{Z}\), which is the same as \cref{eq:dscconv} up to notation. The following example will demonstrate how using \cref{eq:dscconv} in the mathematically justified setting can yield accurate results, how this accuracy breaks down on non-uniformly spaced points, and how our proposed method fixes these issues. To that end, consider the following 1D functions:
            \begin{align*}
                f(x) = \sin(\pi x) + \sin(14 \pi x) & & \text{and} & & g(x) = \frac{8 \sin(8 \pi x)}{\pi x},
            \end{align*}
            where \(f\) is a signal composed of low and high frequency sine waves, and \(g\) is the ideal low-pass filter whose action under convolution will remove the higher frequency from \(f\). In \cref{fig:conv-ex-continuous} we can see the result of this convolution computed analytically according to \cref{eq:ctsconv}.

            \begin{figure}[H]
                \centering
                \begin{subfigure}[b]{\textwidth}  
                    \centering 
                    \includegraphics[width=\textwidth]{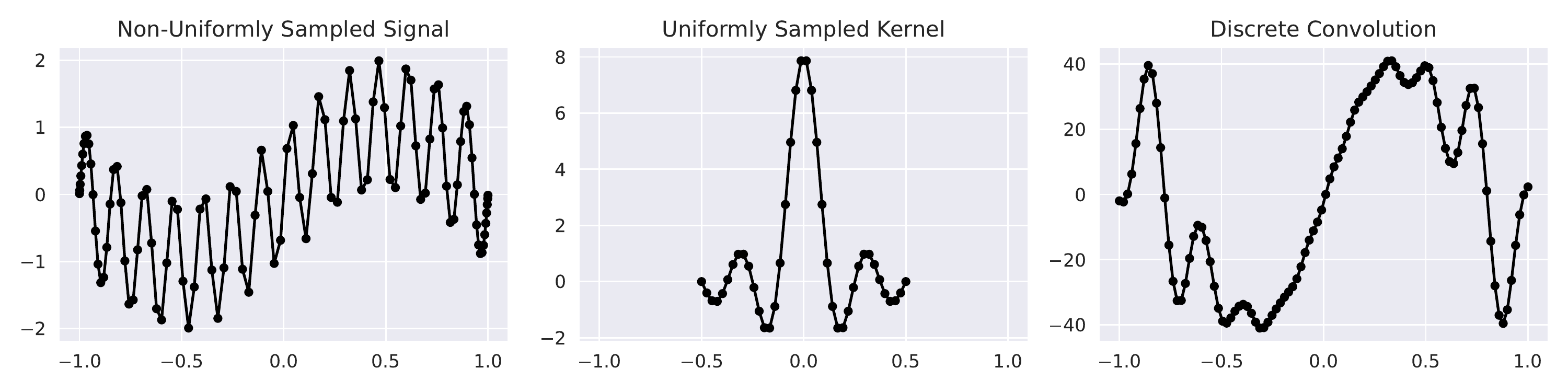}
                    \caption{\small Discrete convolution with uniformly sampled kernel.}
                    \label{fig:conv-ex-mixed-sampling}
                \end{subfigure}
                \vfill
                \begin{subfigure}[b]{\textwidth}  
                    \centering 
                    \includegraphics[width=\textwidth]{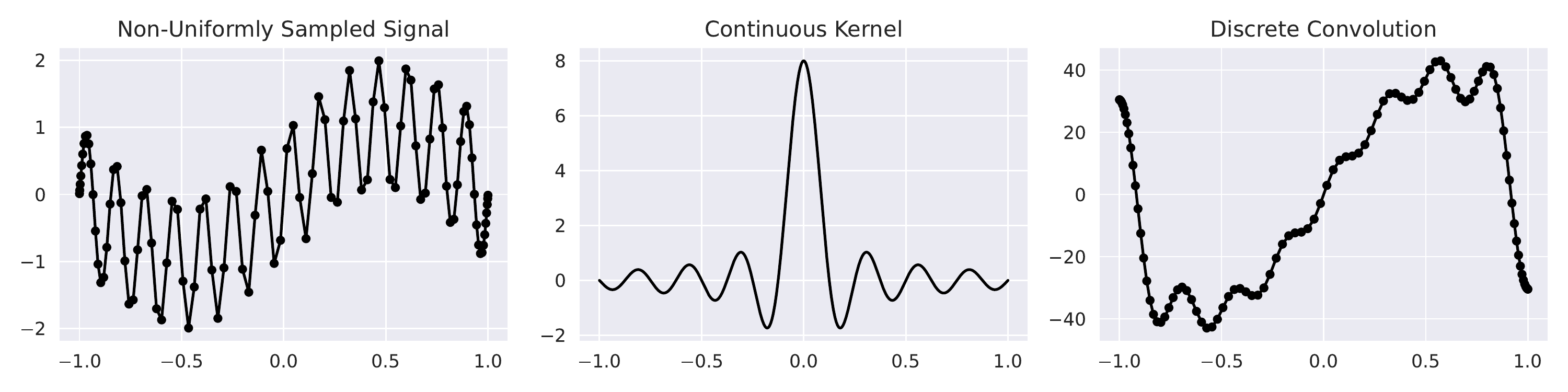}
                    \caption{\small Discrete convolution with continuous kernel.}
                    \label{fig:conv-ex-nonuniform-sampling}
                \end{subfigure}
                \vfill
                \begin{subfigure}[b]{\textwidth}  
                    \centering 
                    \includegraphics[width=\textwidth]{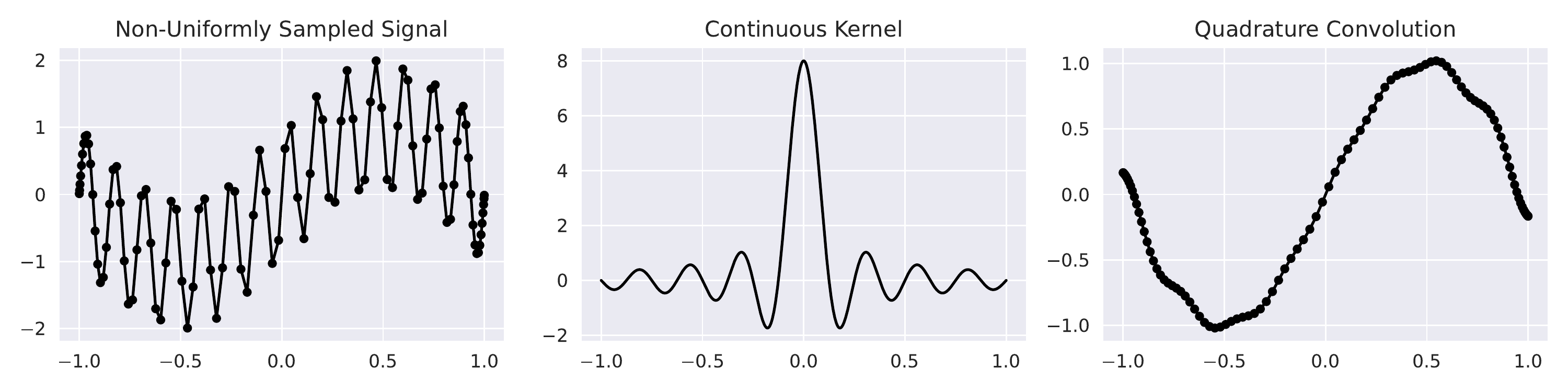}
                    \caption{\small Quadrature-based convolution with continuous kernel.}
                    \label{fig:conv-ex-quadrature}
                \end{subfigure}
                \caption{\small Comparison of discrete convolution and quadrature-based convolution on a non-uniformly sampled signal.} 
                \label{fig:conv-motivation-2}
            \end{figure}
                
            In \cref{fig:conv-ex-uniform-sampling}, one can observe the normal usage of the discrete convolution, \cref{eq:dscconv}, with uniformly spaced grid points, and note that its result closely matches the continuous baseline. \Cref{fig:conv-motivation-2}, on the other hand, considers the results of a number of convolution type operations when the input signal is sampled in a non-uniform manner. \Cref{fig:conv-ex-mixed-sampling} visualizes the results of discrete convolution if we sample the data at non-uniform locations, but continue to use the uniformly sampled kernel. This result only vaguely resembles the analytic output, in part because the kernel isn't sampled at the correct locations. The discrete convolution of \cref{fig:conv-ex-nonuniform-sampling} employs a continuous kernel that can be sampled wherever necessary and performs markedly better. That is to say, the continuous kernel can be evaluated at any point \(y-x_i\), where we have chosen the output points \(y\) to be uniform for visualization purposes. However, the use of a continuous kernel alone is not sufficient to accurately approximate the analytic convolution. Instead, in this work, we propose modeling the kernel, \(g\), as a continuous function and performing the calculation of discrete convolution as a quadrature approximation of \cref{eq:ctsconv}. This yields the results we see in \cref{fig:conv-ex-quadrature}. As we detail in \cref{sec:methods}, the method we propose includes quadrature weights \(\rho_i\) in \cref{eq:dscconv} along with a continuous kernel \(g(\cdot)\), so that our 1D equivalent would be the following:
            \begin{equation}\label{eq:our_method_1d}
                (f\ast g)(y)\approx\sum_i \rho_i\cdot f(x_i)\cdot g(y - x_{i}).
            \end{equation}
            The combination of the continuous kernel and the reduction of integration error gives us the best approximation of the operation we wish to perform. Although this example presumes a fixed continuous kernel, as opposed to a learned kernel, it demonstrates the fundamental problem of naively applying convolutional kernels to discrete data on non-uniform points.
                    
        
        \subsection{Related Work}\label{sec:related-work}

            \paragraph{Graph convolutions}
            Graph convolutions are perhaps the most widely used convolution method for non-uniform data, applicable when the data has a readily available adjacency structure. For a full review of the relevant methods we refer the interested reader to \cite{graphconv}. In general, these methods are either spectral or spatial. Spectral graph convolutions perform convolution in the Fourier domain, computing the convolution as a point-wise product of two signals, whereas spatial graph convolutions work in the spatial domain directly. Graph convolutions do not always have clear connections to standard convolution\edit{; in particular, one cannot down-sample the input directly. Thus, graph convolutional neural networks (GCNNs) rely on pooling methods for down-sampling the number of nodes in an arbitrary graph. Significant effort has been put into graph pooling operators from a variety of perspectives \cite{graphpoolreview}, but note that not all of these have an associated unpooling operator. Many methods remove edges \cite{hanocka2019} or nodes \cite{ranjan2018, gao2019} via a (possibly trainable) score function based on graph features. Other methods create a new down-sampled graph by clustering groups of nodes \cite{suk2022}, where the cluster assignment can be accomplished in numerous ways.}
                        
            Fundamentally, since graph convolutions use only adjacency structure, they do not exploit the full knowledge of non-uniform data.  For example, recall the 1D example of the previous section. In 1D, all grids (uniform or not) of the same number of nodes will have the same adjacency structure, so graph based methods could not distinguish uniform from non-uniform samples. \edit{This makes them 
            less suitable for spatially embedded data, where the coordinates of the points can be used in addition to the graph structure. In \cref{sec:experiments} we will compare against graph convolutions using standard max-pooling by leveraging information outside of the adjacency structure. Effectively, this provides a best case scenario for GCN, and avoids employing more complicated pooling methods.}

            \paragraph{Point cloud convolutions}
            There are numerous approaches to convolution for spatially embedded data, owing to the availability of point cloud data from LiDAR measurements \cite{conv_review}. Some of these methods voxelize the input data and perform convolution on the resulting voxels \cite{voxelize_ex}. This often results in significant sparsity which can be taken advantage of for reasonable computational complexity. However, it eliminates the application to many non-uniform problems, if, for example, maintaining variable point density is desirable, as is commonly done to resolve boundary layer effects in fluid problems. The PointNet architecture \cite{pointnet} pioneered the use of shared Multi-Layer Perceptrons (MLPs) operating directly on the points themselves. Hybrid methods, such as \cite{spvconv}, have shown great success by combining a voxelized convolution branch with a point-based branch. Other methods may perform convolution directly on the points by various methods such as utilizing MLPs to define the convolutional kernels over all the necessary locations \cite{convpoint} or changing the definition of convolution to adapt to the irregular domain \cite{spidercnn,annular_conv}. We believe the most similar of these approaches to our work is PointConv \cite{pointconv}, which attempts to approximate the integral via Monte Carlo integration. However, we approximate the integral via quadrature and since typical meshes are in three dimensions or less, quadrature is typically superior to Monte Carlo. In general, point cloud methods allow for spatial locations to change from sample to sample; in contrast, our method explicitly takes advantage of a static mesh.

            \paragraph{Continuous convolutions}
            \edit{Continuous convolution operators have been investigated for some time, motivated by a variety of use cases including non-uniform input data, arbitrary kernel sizes, and multi-resolution learning. These methods vary widely in their choice of continuous domain, how their kernel is parameterized, and how the operator is constructed. The 2017 work of \cite{edgeconditioned} generates kernels for graph convolution via continuous edge labels. The use of an MLP that maps from coordinates of spatially embedded data to parameterize the kernel was introduced in \cite{wang2018parametric}. This approach is also used in \cite{romero2022general, ckconv, flexconv} to construct arbitrarily sized kernels for efficiently modeling long range dependencies. Other parameterization approaches for the kernel have also been considered. SplineCNN \cite{splinecnn} uses a set of B-Spline basis functions. S4ND \cite{s4nd} constructs a global kernel via a Kronecker product of sampled one-dimensional convolution kernels created as a linear combination of hand-picked basis functions (e.g., sine and cosine). The work of \cite{smpconv} learns both the locations and values of a discretely sampled kernel. The kernel value at an arbitrary query point is then constructed based its local samples. During the preparation of this paper, the work of \cite{ccnn} was made available, which also represents the kernel as an MLP and learns it from data. The authors seek to develop an operation which resembles standard convolutions as much as possible, while still being applicable to non-uniform data. The key difference between the continuous convolution operators discussed so far and our own is the addition of quadrature to weight our finite sum.}

            \paragraph{General applications}
            There are many applications where the density of points is directly related to the underlying data and can be used to achieve more accurate convolution integrals than the standard convolution discretization. Non-uniform meshes are prevalent in PDE simulations, where nodes may be concentrated in areas where the PDE solution has a large gradient, such as near the boundary.

            \paragraph{Compression of scientific data}
            The simulation of PDEs can create immense amounts of data, which then requires compression in order to store on disk for later scientific usage. While more classical methods exist for both lossless \cite{fpzip} and lossy compression \cite{fzp,di2016fast} for these types of datasets, they are largely limited to uniform grids or lose efficiency when applied to non-uniform data. Compression of PDE simulation data with convolutional autoencoders, has proven to be effective on uniform grids achieving high compression ratios (in excess of $100 \times$) with low reconstruction errors \cite{insitu-compression, momenifar2022}. 

            \paragraph{Other scientific applications}
            In addition to data compression, convolutional neural networks have been utilized in PDE super-resolution methods \cite{stengel2020adversarial} and reduced-order modeling \cite{momenifar2021, lee}. All of these techniques could benefit from a generalization of standard convolution to the non-uniform meshes commonly found in PDE simulation data.
        

    \section{Methods}\label{sec:methods}
    
        We will begin by discussing the construction of our discrete convolution operator, and then describe the details associated with a practical implementation. Our approach, QuadConv, will be to approximate \cref{eq:ctsconv} via quadrature, so denote the nodes as \(\bm{x}_i\) and the weights as \(\rho_i\), for \(i=1,\ldots,N\), and then consider the following approximation:
        \begin{equation}\label{eq:quad-scalar}
            (f \ast g)(\bm{y})\approx\sum_{i=1}^N\rho_i\cdot f(\bm{x}_i)\cdot g(\bm{y}-\bm{x}_i).
        \end{equation}
        We employ the following form for our kernel function \(g\):
        \begin{equation}\label{eq:kernel}
            g(\bm{z}) = \bump(\bm{z})\cdot h(\bm{z};\bm{\theta}),
        \end{equation}
        where \(\bm{\theta}\) are learnable parameters for some function \(h:\reals^D\to\reals\). The definition of the bump function is then given below for some \(\alpha>0\):
        \begin{equation}\label{eq:bump}
            \bump(\bm{z})=\begin{cases}\exp{\Big(1-\frac{1}{1-(\|\bm{z}\|/\alpha)^4}\Big)} & \|\bm{z}\|<\alpha \\ 0 & \text{else}\end{cases}.
        \end{equation}
        We observe that \(\alpha\), which is set manually, allows us to control the support of \(g\), and therefore the effective size of the kernel. As discussed earlier, this compact support is desirable because we seek to extract local features from the input data. It also has the important consequence of significantly reducing the computational overhead, which we will see in more detail in \cref{sec:computation}. \edit{When setting $\alpha$ in practice, we can take the same approach as standard discrete convolutional layers where $3 \times 3$ kernels are often preferred in 2-dimensions since multiple layers with smaller kernels can replicate the effect of a larger kernel size. Similarly, we prefer smaller compact supports to larger ones which saves computation and allows us the same flexibility when layers are composed. When the domain is non-uniform we set $\alpha$ based on the average number of points inside the compact support.}
        
        We have written equation \cref{eq:quad-scalar} as a single sum based on the direct quadrature approximation of \cref{eq:ctsconv}, but this can also be viewed as an iterated integral and a corresponding iterated sum. For example, in two dimensions we could let \(\bm{x}=(x_1,x_2)\) and \(\bm{y}=(y_1,y_2)\) to rewrite \cref{eq:quad-scalar} as:
        \begin{equation}
            (f \ast g)(y_1, y_2)\approx\sum_{i=1}^{N_1}\sum_{j=1}^{N_2}\rho_{ij}\cdot f(x_{1_i},x_{2_j})\cdot g(y_{1}-x_{1_i},y_{2}-x_{2_j}),
        \end{equation}
        where we have quadrature weights \(\rho_{ij}\) for \(i=1,\ldots,N_1\) and \(j=1,\ldots,N_2\). In fact, if the points lie on a uniform grid, then using the two-point composite Newton-Cotes quadrature recovers standard convolution. Such a choice of quadrature yields weights \(\rho_{ij}=1/4\), and equates to using the trapezoidal rule along each dimension. This constant weight can then be absorbed into the learned kernel, producing the simple finite sum we see with standard convolutions.

        The specifics of the quadrature (i.e., the nodes and weights) is a question so far unaddressed. In general, we consider the nodes to be given as part of the input data. In other words, the quadrature nodes are fixed by the locations at which the input signal \(f\) is evaluated. In many scenarios the weights can be readily computed on the fly. For example, if the input points are on a uniform grid, then the operator may use Newton-Cotes weights. For data from FEM based simulations it is possible to construct node based quadrature weights from individual element quadrature evaluations, but this is a somewhat complicated procedure that could incur significant computational overhead 
        \edit{and furthermore the usual quadrature construction depends on the kernel}. 
        Because we already expect to learn the kernel (i.e., \(\bm{\theta}\)) from data, perhaps the easiest option is to learn the quadrature weights as well. When we learn the quadrature weights we ensure they are strictly positive in order to avoid catastrophic cancellation and resulting loss of precision \cite{float_handbook}.
        
\begin{figure}[h]
            \centering
            \begin{subfigure}[b]{0.475\textwidth}
                \centering
                \includegraphics[width=\textwidth]{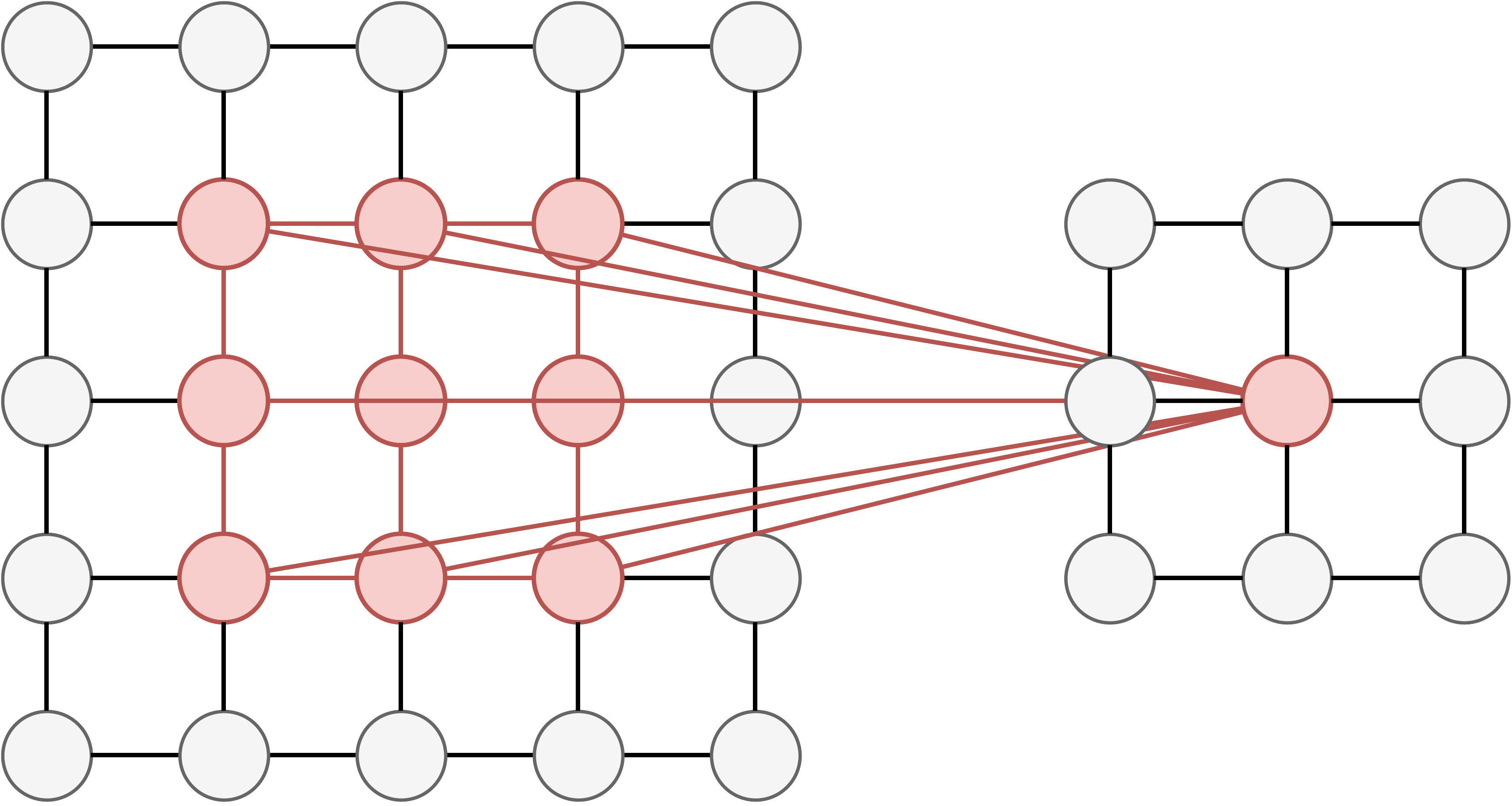}
                \caption{\small Discrete convolution with \edit{3x3 kernel} on uniform grid.} 
                \label{fig:standard-conv}
            \end{subfigure}
            \hfill
            \begin{subfigure}[b]{0.475\textwidth}  
                \centering 
                \includegraphics[width=\textwidth]{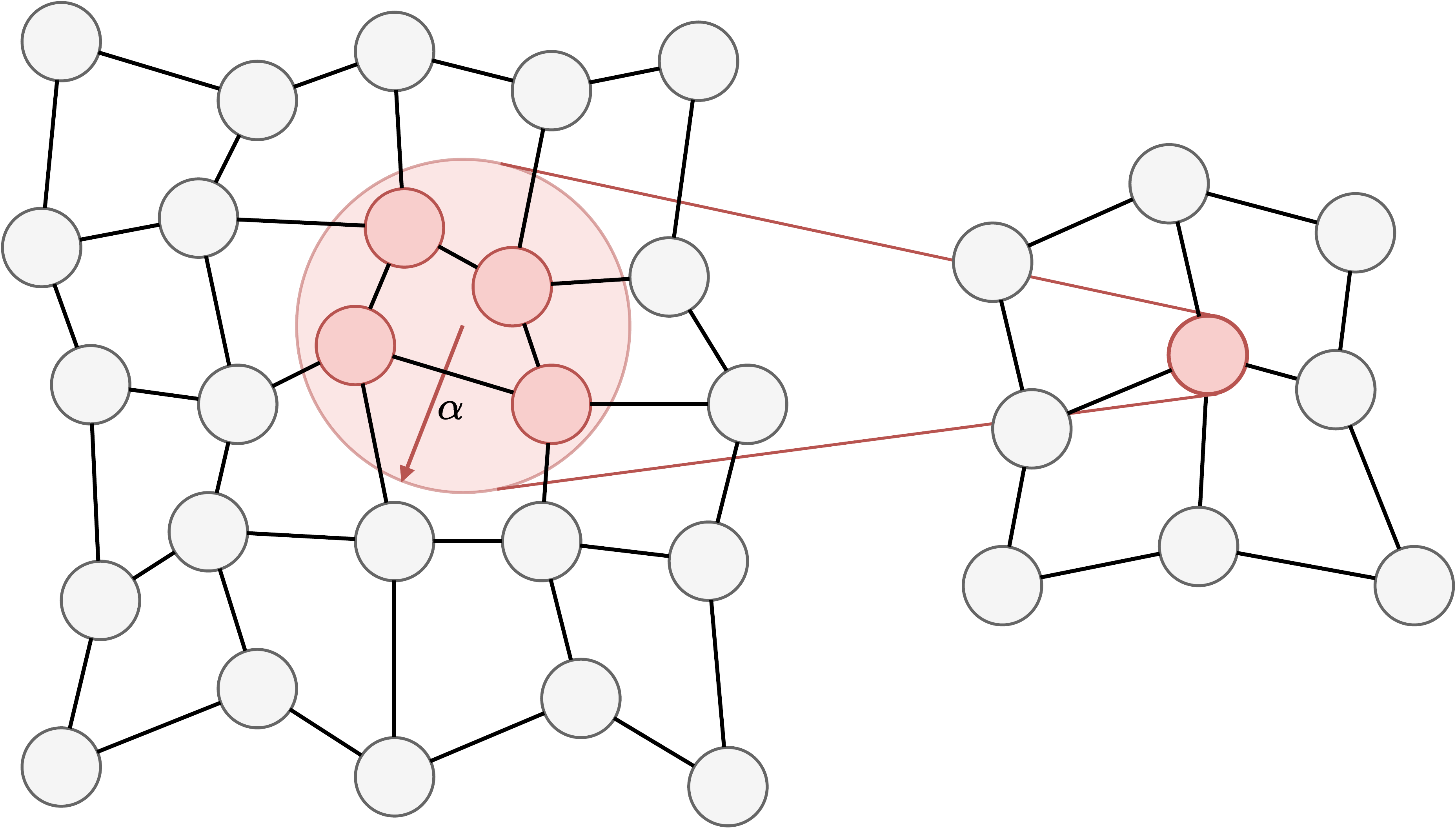}
                \caption{\small QuadConv with \edit{\(\alpha\) compact support} on non-uniform mesh.}
                \label{fig:quadrature-conv}
            \end{subfigure}
            \caption{\small Comparison of discrete convolution and QuadConv.} 
            \label{fig:conv-comparison}
        \end{figure}

        \Cref{fig:conv-comparison} visualizes, in two dimensions, the difference between the standard discrete convolution and our quadrature convolution. The standard method uses a grid based kernel ($3\times3$ in this case) that slides across the spatial dimensions of the domain to compute the output values. However, when one moves to a non-uniform mesh, this operation no longer applies. On the other hand, the quadrature method computes the output values using mesh nodes that lie within the compact support (translucent red circle) of the kernel, and so it is agnostic to the underlying mesh structure.
        
        Analytic convolutions (\cref{eq:ctsconv}) are equivariant to continuous translations, and their standard discretization (\cref{eq:dscconv}) inherits this property in a discrete sense. In mathematical terms, if \(T\) is some translation operator \((Tf)(\bm{x})=f(\bm{x}-\bm{t})\), then the following holds:
        \[T(f \ast g)=(Tf) \ast g = f \ast (Tg),\]
        or, more plainly, the output of the convolution is translated equivalently to the translation of its input. In image classification, for example, this property is desirable because features should be invariant to placement in the image.
%
%
        Because QuadConv approximates continuous convolution, it is approximately equivariant, up to the accuracy of the quadrature.
        As features move through the domain, the output of the QuadConv layers will vary by approximately the same amount as the original data.
        
        So far, we have only considered scalar valued inputs to the convolution operator, but it is necessary to extend this to a multi-channel (i.e., vector) setting. Standard deep learning convolutions accomplish this by stacking multiple kernels into a filter, and then applying multiple filters to the input data. We will adopt a similar approach. Define \(\bm{G}:\reals^D\to\reals^{\tilde{C}\times C}\) as a map from a point to a matrix, which we interpret as a stack of filters. Despite the incongruence in vocabulary, we will refer to \(\bm{G}\) as a filter. Now, we consider \(\bm{f}:\reals^D\to\reals^{C}\) as our vector valued input, and the quadrature convolution is given as:
        \begin{equation}\label{eq:quad-vector}
            (\bm{f}\ast\bm{G})(\bm{y})=\sum_{i=1}^N \rho_i\cdot \bm{G}(\bm{y}-\bm{x}_i)\cdot\bm{f}(\bm{x}_i),
        \end{equation}
        where we can easily observe that the output lies in \(\reals^{\tilde{C}}\). The filter \(\bm{G}\) is parameterized by \(\bm{\theta}\) according to the multi-channel extension of \cref{eq:kernel}:
        \begin{equation}\label{eq:filter}
            \bm{G}(\bm{z})=\bump(\bm{z})\cdot\bm{H}(\bm{z};\bm{\theta}),
        \end{equation}
        where \(\bm{H}(\cdot;\bm{\theta}):\reals^D\to\reals^{\tilde{C}\times C}\). The form of the bump function remains unchanged, and thus we see that \(\bm{G}\) may be close to \(\bm{0}\) depending on the location it is evaluated at. We may also note that the dimension, \(D\), of the data was arbitrary, so our operator can easily be applied to data in any dimension. 
        
        
        \subsection{Practical Computation}\label{sec:computation}
        
            The methodology we have developed so far is the foundation of our proposed convolution operator, but certain aspects of the implementation are as of yet unclear. Further, specific elements of the computation must be leveraged to achieve any reasonable level of efficiency. We will discuss the practical implementation of the quadrature-based convolution operator, and consider its asymptotic computational complexity. 
            
            In general, we consider input in the form of a tuple \((\bm{X},\bm{F})\), where \(\bm{X}\in\reals^{D\times N}\) represents the mesh consisting of \(N\) points in \(D\)-dimensional space, and \(\bm{F}\in\reals^{C\times N}\) are the associated features with \(C\) channels. For example, a 3D flow field would have three channels; one each for the velocity along each dimension. For a training dataset with $T$ training samples, the input will be $\left( (\bm{X},\bm{F}^t) \right)_{t=1}^T$, meaning that all samples share the same mesh. \edit{For a practical implementation of the current version of QuadConv we require a single fixed input mesh. If we were to allow the mesh to vary, it would significantly increase the cost of learning quadrature weights for each node, and negate the speedups obtained via caching, which is discussed later in this section.}
            
            The QuadConv operator, which we will denote as \(\bm{Q}\), acts in the following manner:
            \begin{equation*}
                (\bm{X},\bm{F})\xrightarrow{\bm{Q}}(\bm{Y},\bm{\tilde{F}}),
            \end{equation*}
            where \(\bm{Y}\in\reals^{D\times \tilde{N}}\) are \(\tilde{N}\) output points, and \(\bm{\tilde{F}}\in\reals^{\tilde{C}\times \tilde{N}}\) are the recovered features with \(\tilde{C}\) channels. As with standard convolutions, there are a number of architectural decisions that must be made manually by the user, and each comes with a variety of trade-offs. Similar to standard convolutions, the practitioner is left to specify the number of output channels. Unlike standard convolutions, where the number of output points is determined as a combination of various hyperparameters such as stride and padding, the number of output points and their coordinates are directly specified when using QuadConv. Due to this, our method easily facilitates up-sampling or down-sampling, i.e., increasing or decreasing the number of points in the domain, and can maintain non-uniform point density. This is due to the continuous kernel being defined at all points inside the domain which maximizes our re-sampling flexibility --- this means there are no inherent constraints on the output points of a QuadConv layer. Thus the process by which the output point coordinates are computed is an important decision factor. In the simplest case, assuming the geometry is Cartesian, they can be placed on a uniform grid. A more sophisticated strategy would involve coarsening the underlying mesh via agglomeration, e.g., \cite{mesh_agglomeration}, to construct the output points directly from the input points. The actual computation by \(\bm{Q}\) via \cref{eq:quad-vector} is given below:
            \begin{equation}\label{eq:qc-computation}
                \bm{Q}[\bm{X}, \bm{F}]_j=(\bm{Y},\bm{\tilde{F}})_j=\big(\bm{y}_j,\sum_{i=1}^N \rho_i\cdot \bm{G}(\bm{y}_j-\bm{x}_i)\cdot \bm{f}_i\big),
            \end{equation}
            for \(j=1,\ldots,\tilde{N}\). Note that \(\bm{y}_j\) and \(\bm{f}_i\) are the \(j\)\textsuperscript{th} and \(i\)\textsuperscript{th} columns of \(\bm{Y}\) and \(\bm{F}\), respectively. Recall from \cref{eq:filter} that \(\bm{H}\) provides us with the learnable map from a point to a matrix. Our method, in general, is agnostic to the form of this map, but for the sake of efficiency we opt for a single MLP, $\bf{G}: \mathbb{R}^D \rightarrow \reals^{\tilde{C} \times C}$. There are many other possible choices, such as using multiple MLPs which are shared among input/output channels or are entirely independent. 
            
\begin{figure}[h]
                \centering
                \includegraphics[width=\textwidth]{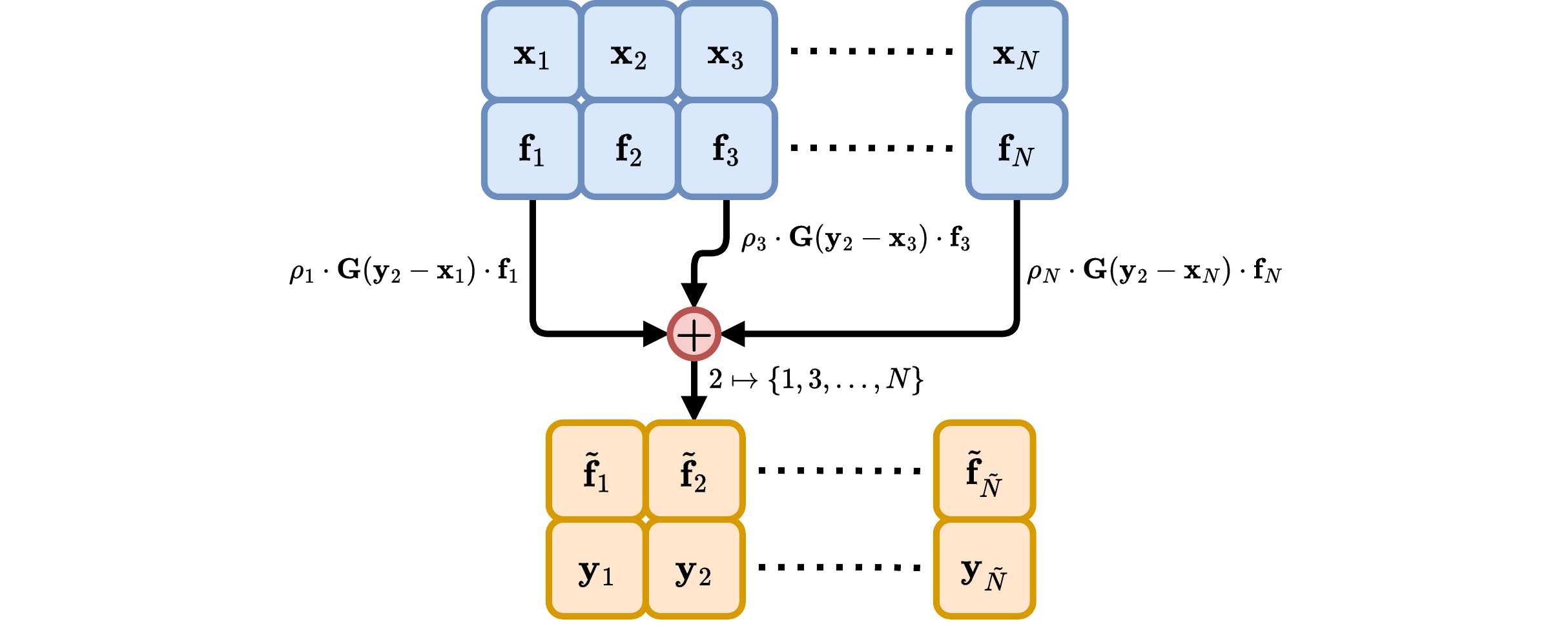}
                \caption{\small QuadConv computation. The output value at index \(j=2\) depends only sparsely on the input values, e.g., there is no dependence on the input value at index \(i=2\).} 
                \label{fig:qc-operator}
            \end{figure}

            As noted earlier, depending on their evaluation location, the approximation \(\bm{G}\approx \bm{0}\) may hold for many of the filters, allowing the implementation to take advantage of the sparsity of the sum in \cref{eq:quad-vector} in order to avoid undue computation. The set of index pairs $i$ and $j$ for which the function \(\bump(\bm{y}_j-\bm{x}_i)\) is nonzero generates a map of the form \(j\mapsto\{i_k\}\), meaning that for each output location index $j$, there is a set of input indices \(\{i_k\}\) that contribute to its value. \Cref{fig:qc-operator} visualizes how this map can be used to implement \cref{eq:qc-computation}.
            
            This entire process can be executed efficiently in a vectorized manner as the individual operations are independent, and for any given output index \(j\) the sum that determines its features is over far fewer indices \(i\) than would otherwise be the case, just as in the standard discrete case. Assuming that the input and output locations are static, then the maps \(j\mapsto\{i_k\}\) can be computed once as a pre-processing step and cached for future use. Doing so avoids the costly construction of the maps themselves, which significantly improves the execution time of the convolution operator.
             
            \Cref{tab:conv-complexity} shows the computational complexity for the forward pass of QuadConv compared to standard discrete convolutions. For QuadConv, we split the complexity into the construction step for the index map, which needs only be computed once for the entire dataset and hence is amortized, and the actual quadrature convolution computation. This helps to highlight the impact of caching on the overall performance. 
            During training, there is also cost due to the backward pass in automatic differentiation, which we assume is on the same scale as the cost of forward pass.
            
            The variables \(\{C,N\}_{in}\) and \(\{C,N\}_{out}\) are the input and output channels and points, respectively. We consider a standard convolution with kernel size \(K\), and for QuadConv, we consider a sparsity factor \(S\), \edit{the average number of nodes inside the support of the kernel $\bm{G}$}, which is dependent on the 
            \edit{radius $\alpha$}
            of the bump function and the local point density. The variables \(M_t\) and \(M_m\) denote the time and memory complexity for the chosen filter operation (i.e., MLP). Note that \(M_m\) has an implicit weak dependence on \(C_{in}\) and \(C_{out}\), as the output of the filter operation is some \(C_{in}\times C_{out}\) matrix.
            \begin{table}[h]
                \centering
                \begin{tabular}{lccc}
                    \toprule
                    & \textbf{Standard} & \multicolumn{2}{c}{\textbf{QuadConv}} \\
                    \cmidrule(l){3-4}
                    & & \textbf{Convolution} & \textbf{Index Map} \\ 
                    \midrule
                    \textbf{Time} & \(K N_{out}C_{in}C_{out}\) & \( \edit{SN_{out}C_{in}C_{out} + M_{t}SN_{out}}\) &\(N_{in}N_{out}\) \\
                    & \(KNC^2\) & \(\edit{SNC^2 + M_tSN}\) & \(N^2\) \\
                    \midrule
                    \textbf{Peak Memory} & \(K C_{in}C_{out}\) & \(M_m + C_{in}C_{out}\) & \(SN_{out}\) \\
                    & \(KC^2\) & \(M_m + C^2\) & \(SN\) \\
                    \bottomrule
                \end{tabular}
                \caption{Big-\(\mathcal{O}\) comparison of standard discrete convolutions and QuadConv. The second row of the time and memory blocks takes \(N_{in}=N_{out}=\mathcal{O}(N)\), and \(C_{in}=C_{out} = \mathcal{O}(C)\) in order to facilitate comparisons. The index map only needs to be computed once for a given mesh.}
                \label{tab:conv-complexity}
            \end{table}
            When considering the peak memory consumption, we do not include the requirements for the input or output data, as that is common to all three methods. 
            
            One may observe a number of important details from \cref{tab:conv-complexity}. With respect to QuadConv, we see that sparsity is extremely important. \edit{Just as in standard discrete convolution the local support, or sparsity, of the operator reduces the complexity to \(SN_{out}\) for a single input/output channel, where typically $S\lesssim 10$, down from \(N_{in}N_{out}\) without exploiting sparsity.} From the last column, we can see that because the index map construction must look at all input-output point pairs, it is an \(\mathcal{O}(N^2)\) operation --- producing the \(SN_{out}\) indices needed for the forward pass of QuadConv. This is the fundamental trade-off: caching as a pre-processing step saves significant computation time in return for a larger memory footprint.
            
            
    \section{Numerical Experiments}\label{sec:experiments}

        We have implemented QuadConv using PyTorch \cite{pytorch} and Lightning \cite{lightning}, available as an open source Github repository \cite{quadconv-repo}. The code \edit{and datasets} directly associated with this paper, and capable of recreating our results, can be found in \cite{paper-repo}. This section shows a number of experiments that validate our method of convolution as an effective tool in deep learning. In particular, we perform data compression of PDE simulation data using an autoencoding neural network. First, we start with data on a uniform grid (\cref{sec:uniform-ignition}) so that we can establish a baseline of performance by comparing our method to standard discrete convolution. We then interpolate this dataset onto a non-uniform mesh and show that QuadConv still performs just as accurately as with the uniform mesh (\cref{sec:non-uniform-ignition}). Finally, we consider a different dataset which was simulated on a non-uniform mesh, and show that our operator continues to facilitate high reconstruction accuracy at large levels of compression (\cref{sec:non-uniform-flow}).
        
\begin{figure}[h]
            \centering
            \includegraphics[width=\textwidth]{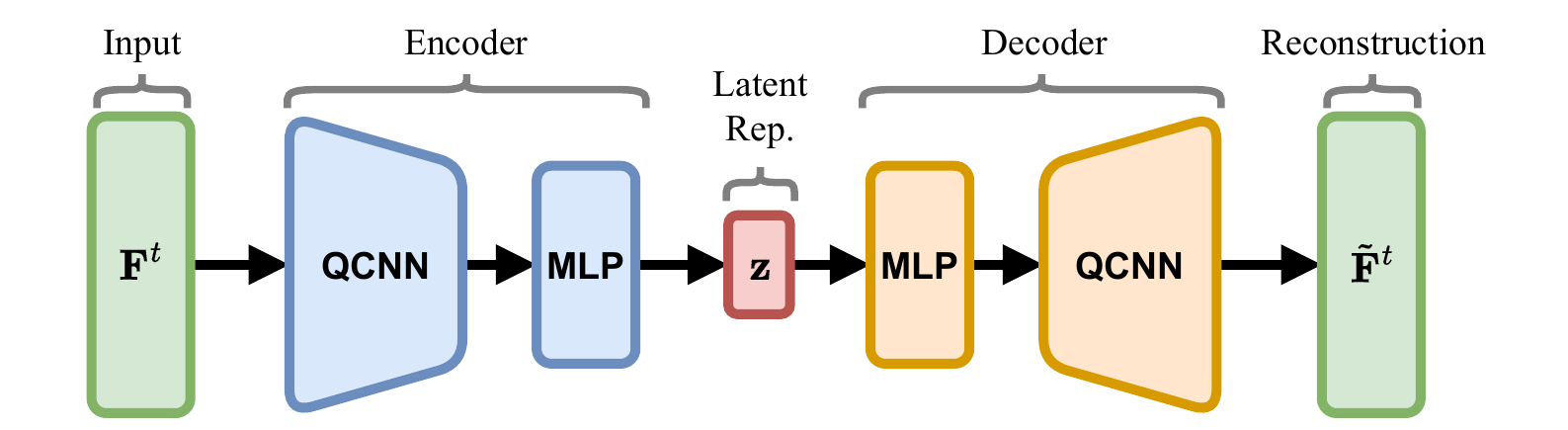}
            \caption{\small Autoencoder structure.}
            \label{fig:autoencoder}
        \end{figure}

        \Cref{fig:autoencoder} provides a high-level visualization of our neural network architecture. The raw input data is passed through the encoder, which is comprised of a (Quadrature) Convolutional Neural Network ((Q)CNN) that progressively down-samples the number of points and increases the number of channels, and an MLP which outputs the latent representation \(\bm{z}\). This latent vector is the compressed state of the data. To reconstruct, one simply performs the reverse operation using the decoder. An MLP is applied to the latent state \(\bm{z}\), and then a (Q)CNN up-samples the data.

        \edit{We use a randomized non-contiguous 80/20 split of our data into a training and testing sets. For data compression where all of the data is available at the time of compression, all of the data is available for training and generalization is not important. In more typical machine learning applications the generalization of the model is the most important metric. We evaluate our metrics over the entire dataset (both testing and training), and we also include the maximum error over the entire dataset to capture the generalization of the operators.}
        
        Our autoencoder aims to minimize the empirical risk of a regularized mean-squared error loss between \(\bm{F}^t\) and \(\bm{\tilde{F}}^t\),
        \begin{equation}\label{eq:loss}
            \mathcal{L}(\bm{F}^t,\bm{\tilde{F}}^t) = \| \bm{F}^t - \bm{\tilde{F}}^t \|_{\fro}^{2} + \lambda R(\bm{F}^t,\bm{\tilde{F}}^t),
        \end{equation}
        where \(t=1,\ldots,T\) indexes the time dimension of the (time-dependent) PDE simulation and \(\|\bm{F}^t\|_{\fro}=\big(\sum_{ij}|\bm{F}^t_{ij}|^2\big)^{1/2}\) is the Hilbert-Schmidt (aka Frobenius) norm. Our regularization term $R(\bm{F}^t,\bm{\tilde{F}}^t)$ in the uniform grid case is the mean-squared error between finite difference derivatives of \(\bm{F}^t\) and \(\bm{\tilde{F}}^t\) (i.e., the Sobolev norm). In our non-uniform examples the derivatives are more difficult to define so we do not use any regularization. The final error we report is the time-averaged relative Frobenius-norm,
        \begin{equation}\label{eq:relative-reconstruction-error}
            \frac{1}{T}\sum_{t=1}^T\frac{\|\bm{\tilde{F}}^t-\bm{F}^t\|_\fro}{\|\bm{F}^t\|_\fro}.
        \end{equation}
        Our data is represented with 32-bit floating point numbers (single precision) \cite{single_precision}, and all training and evaluation was performed with single precision numbers. For the purposes of this discussion, we define the compression ratio as the ratio between the dimension of the raw and compressed data. Since we only compress the data in spatial dimensions, this corresponds to the size reduction of any individual spatial sample. For example, raw data on a \(10\times 10\) grid compressed to a latent space in \(\reals^{10}\) would correspond to a compression ratio of 10. We do not consider the storage of the decoder itself, as this has negligible contribution to the compression ratio as $T\to\infty$; \edit{as we'll show in \cref{tab:uniform-ignition-timing}, storage of the encoder and decoder together is just a few megabytes.}
        
        All experiments were conducted on 4 V100 GPUs and 1 IBM Power9 CPU with the Adam \cite{adam_opt} optimizer. \edit{Nearly all experiments used a batch size of 8 under a distributed data parallel training strategy, except for the SplineCNN networks which use a batch size of 1. The SplineCNN steps are noted with an asterisk in each of the tables that follow but the other experiments each correspond to the same amount of data processed per step.}
        
        
        \subsection{Uniform Grid Ignition Data}\label{sec:uniform-ignition}
        
\begin{figure}[b]
                \centering
                \includegraphics[width=\textwidth]{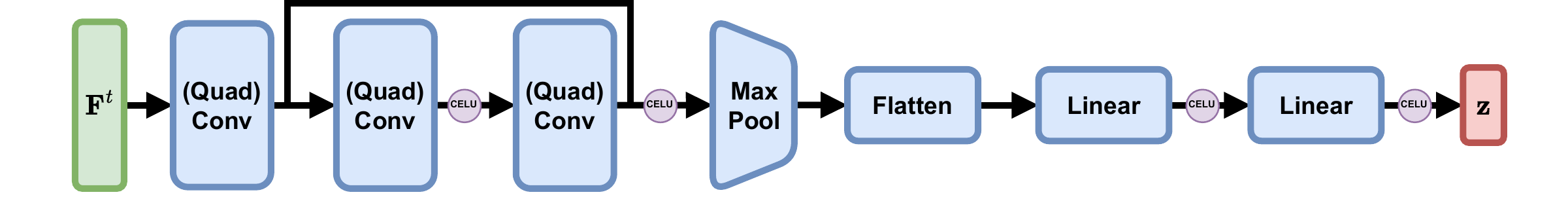}
                \caption{\small Max-pooling based encoder structure. The linear layers comprise the MLP. The decoder is similarly structured, but mirrored to up-sample.}
                \label{fig:pool-encoder}
            \end{figure}

            This dataset consists of 450 uniformly sampled time steps of a jet ignition simulation that lie on a uniform spatial grid of size $50 \times 50$. The wavefront is fully resolved in time. This dataset is transport dominated for the initial portion of the simulation until the jet flame reaches steady state. \Cref{fig:pool-encoder}, with a latent state \(\bm{z}\in\reals^{50}\), describes the model architecture used on the dataset in question --- resulting in a compression ratio of \(50\). Because the input points are already on a uniform grid, we can use the two-point composite Newton-Cotes quadrature \cite{atkinson} weights for the QuadConv layers. This matches the quadrature interpretation of the CNN which gives a very direct comparison of the QuadConv layer to the standard convolution layer. We refer to that QuadConv experiment as "Static Weights" in \cref{tab:uniform-ignition-results}. We also perform the same experiment with quadrature weights as parameters of the neural network, with the hope that the training process can learn a quadrature which is more tailored to the dataset (the "Learned Weights" experiment). In our neural network, the max pooling layer handles the downsampling, so the input and output points for the QuadConv layers are identical.

            \edit{In addition we compare to a few alternatives: a Proper Orthogonal Decomposition (POD) over 50 basis vectors \cite{properorthodecomp}, a graph convolutional network (GCN), and a SplineCNN \cite{splinecnn} network. We choose to use the particular graph convolution operation presented in \cite{kipf2016semi}, which is an approximation of the full spectral graph convolution operator. In practice, many implementations of graph neural networks exhibit instabilities and large maximum errors, the latter of which we observe in the \cref{tab:uniform-ignition-results}. 
            SplineCNN utilizes a continuous kernel which operates on the attributes of the edges given in pseudo-coordinates as well as the adjacency structure of the data. In both examples we circumvent the complexity of graph pooling operators and make the comparison more direct by using the max pooling operator as defined on a grid as the pooling operator.  
            Both the GCN and SplineCNN share the same structure as \Cref{fig:autoencoder,fig:pool-encoder}, also with \(\bm{z}\in\reals^{50}\).}
            \begin{table}[h]
                \centering
                \begin{adjustbox}{max width=\textwidth}
                    \begin{tabular}{lccccc}
\toprule
\textbf{Model Type} & \textbf{Average Error} & \textbf{Max Error} & \textbf{Training Time (h)} & \textbf{\# of Steps} & \textbf{\# of Trainable Parameters} \\ 
\midrule
\textbf{POD (50 Basis Vectors)} & 1.37\% & 17.3\% & N/A & N/A & N/A \\
\textbf{Conv} & 0.42\% & 10.7\% & \hphantom{1}1.46 & 186,000 & 1,024,718 \\
\textbf{GraphConv} & 0.52\% & \hphantom{1}9.1\% & \hphantom{1}2.64 & 315,360 & 1,016,206 \\
\textbf{SplineCNN} & 0.65\% & 10.7\% & 13.85 & 2,728,080* & 1,025,710\\
\textbf{QuadConv (Static Weights)} & 0.63\% & \hphantom{1}\textbf{2.3\%} & \hphantom{1}3.91 & 186,000 & 1,035,310 \\
\textbf{QuadConv (Learned Weights)} & 0.49\% & \hphantom{1}\textbf{1.2\%} & \hphantom{1}9.25 & 290,243 & 1,040,310 \\
\bottomrule
                    \end{tabular}
                \end{adjustbox}
                \caption{\small Autoencoder results for uniform grid ignition data compression at $50\times$ compression.}
                \label{tab:uniform-ignition-results}
            \end{table}

            \begin{table}[h]
                \centering
                \begin{adjustbox}{max width=\textwidth}
                    \begin{tabular}{lccccc}
                        \toprule
                        \textbf{ } & \textbf{Conv} & \textbf{GraphConv} & \textbf{SplineCNN} & \textbf{QuadConv (Static)} & \textbf{QuadConv (Learned Weights)} \\ 
                        \midrule
                        \textbf{Model Size (MB)} & 3.98 & 3.95 & 4.88 & 4.08 | 13.73 (Cached) & 4.08 | 13.73 (Cached)  \\
                        \textbf{Inference Time (ms)} & 2.83 & 8.98 & 4.86 & 8.63 & 8.55 \\
                        \bottomrule
                    \end{tabular}
                \end{adjustbox}
                \caption{\small Inference time and model size for all autoencoders on a uniform grid.}
                \label{tab:uniform-ignition-timing}
            \end{table}
            %
    
\begin{figure}[H]
                \centering
                \begin{subfigure}[b]{7.9cm}
                    \centering
                    \includegraphics[clip,width=\textwidth]{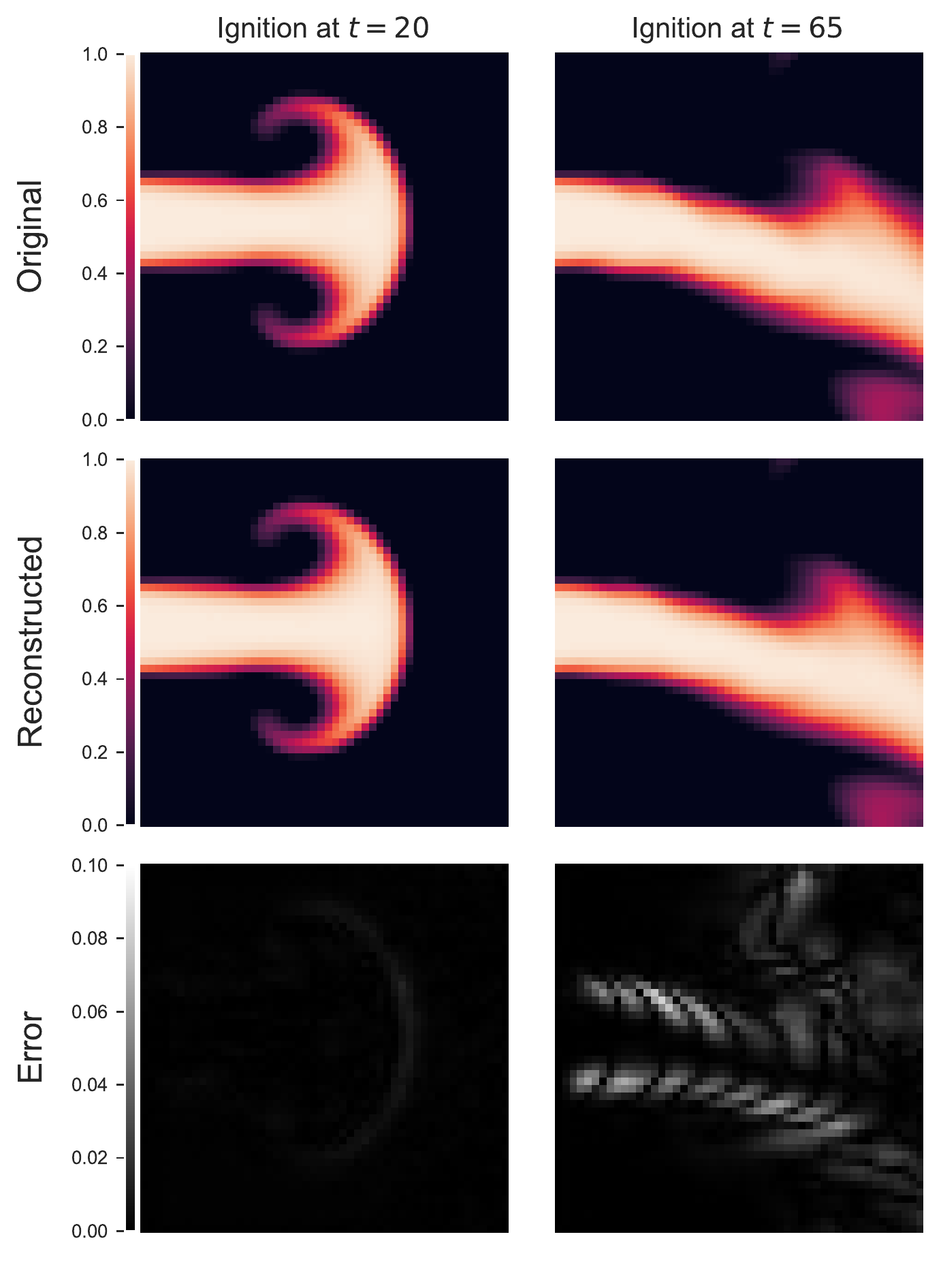}
                    \caption{\small CNN compression.} 
                    \label{fig:cnn-uniform-viz}
                \end{subfigure}
                \hfill
                \begin{subfigure}[b]{7.458cm}
                    \centering 
                    \includegraphics[clip,width=\textwidth]{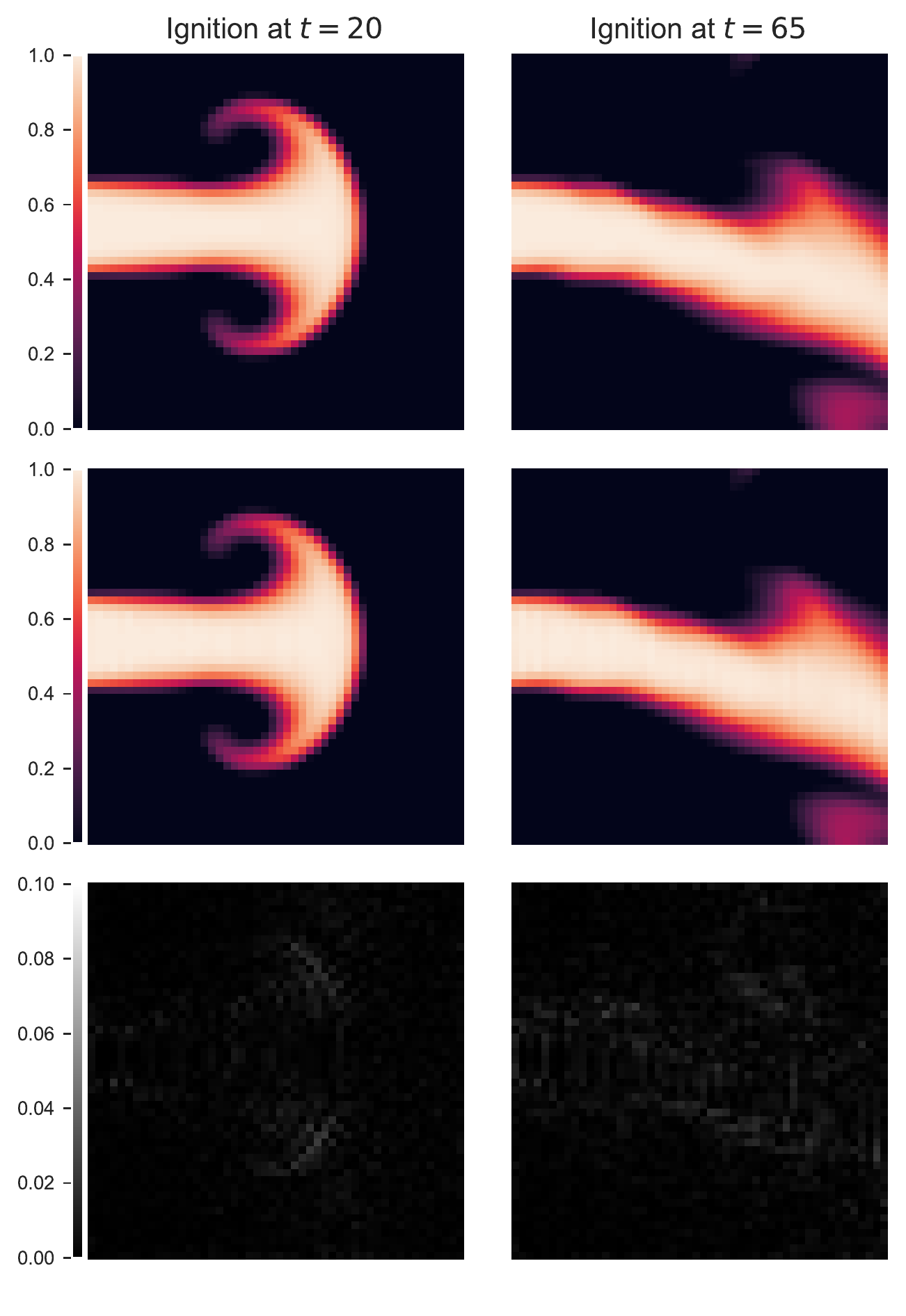}
                    \caption{\small QCNN compression.}
                    \label{fig:qcnn-uniform-viz}
                \end{subfigure}
                \caption{\small Comparison of CNN and QCNN ignition data reconstruction. Note that the bottom row has a rescaled color bar.}
                \label{fig:uniform-ignition-reconstruction}
            \end{figure}

            We can see in \cref{tab:uniform-ignition-results} that the CNN and QCNN indeed have very similar average performance, although the QCNNs take longer to train. Learning the quadrature weights provides a slight advantage for the \edit{average} reconstruction error \edit{and a large improvement in the maximum error}, though it comes at the cost of a longer training time to converge relative to the static quadrature.  \edit{We see that the GCN has competitive average reconstruction error but the worst maximum error.} While increased training time when using QuadConv is to be expected, at roughly 2.7$\times$ for the static quadrature, it is not prohibitively more costly. Furthermore, the implementation of standard discrete convolution in PyTorch \cite{pytorch} is exceptionally well optimized, while QuadConv has room for improvement. Even at these relatively high compression ratios, we see few readily visible errors in the data in \cref{fig:uniform-ignition-reconstruction} at selected time points. The error visualization at the bottom compresses the color scale by an order of magnitude in order to better expose the error that is present.
            
            \begin{figure}[H]
                \centering
                \includegraphics[width=\textwidth]{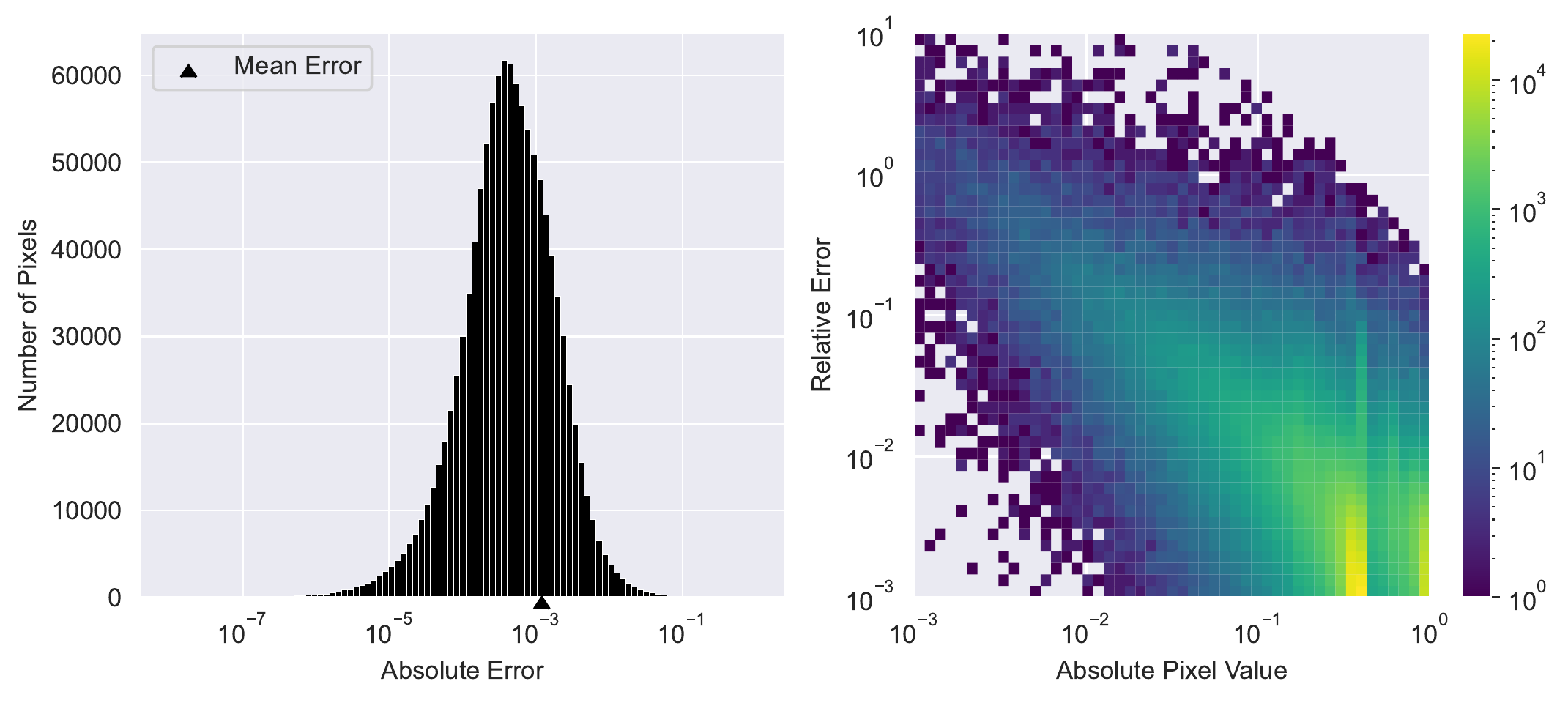}
                \caption{\small Error analysis of CNN for the uniform grid ignition data.} 
                \label{fig:cnn-error}
            \end{figure}

            \begin{figure}[H]
                \centering 
                \includegraphics[width=\textwidth]{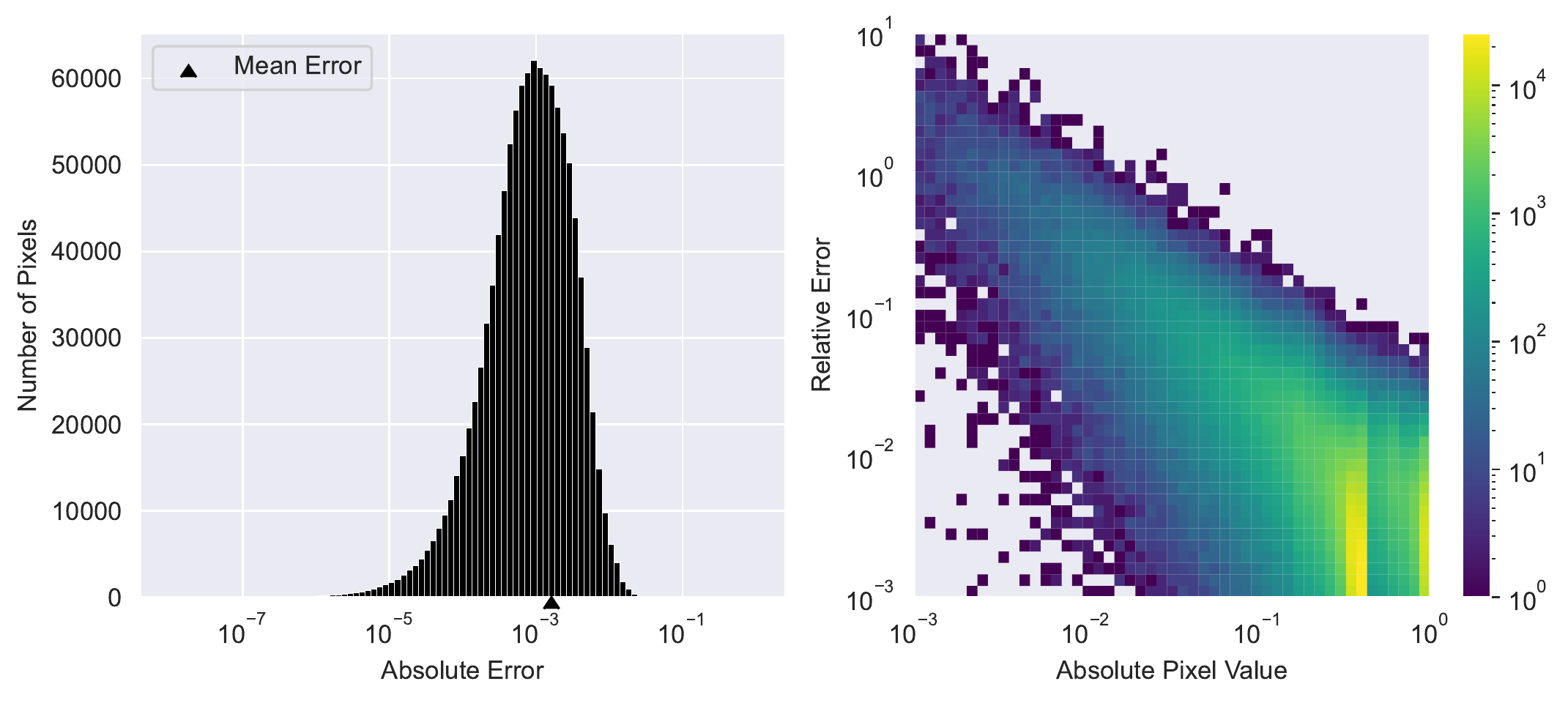}
                \caption{\small Error analysis of QCNN with learned quadrature weights for the uniform grid ignition data.}
                \label{fig:qcnn-error}
            \end{figure}

            We conduct a further error analysis to understand relative performance in \cref{fig:cnn-error,fig:qcnn-error}, which more clearly shows the similarity of performance between the \edit{CNN and the QCNN}. This shows the distribution of the absolute errors are similar, as well as the relative error histogram, which helps to identify at what absolute pixel values large relative errors occur. \edit{The GCN \cref{fig:gcn-error} and SplineCNN \cref{fig:splinecnn-grid-error} have a similar mean absolute error to the CNN and the QCNN but the relative error histograms reveals the larger maximum errors, which can be found in the appendix. }
            

        \subsection{Non-Uniform Mesh Ignition Data}\label{sec:non-uniform-ignition}
        
\begin{figure}[t]
                \centering
                \begin{subfigure}[c]{0.475\textwidth}  
                    \centering
                    \includegraphics[width=0.6\textwidth]{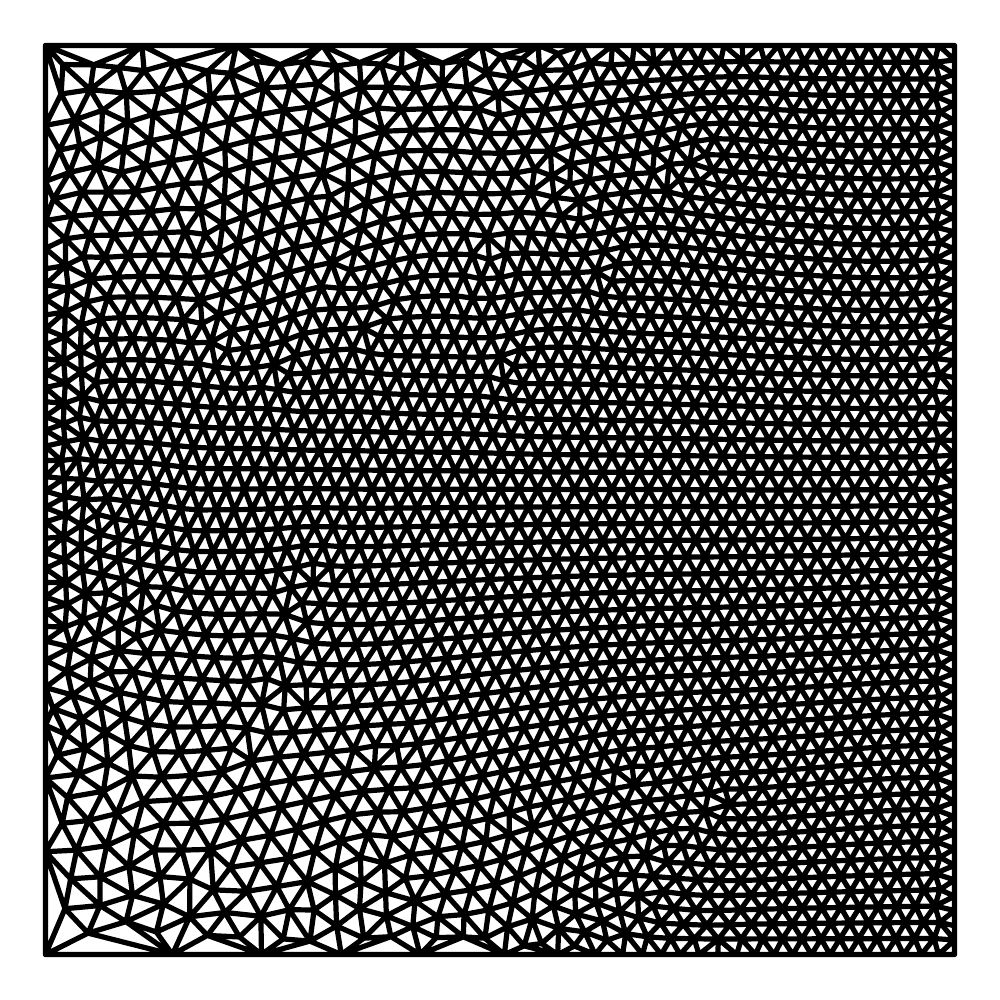}
                    \caption{\small Non-uniform ignition mesh.} 
                    \label{fig:ignition_mesh}
                \end{subfigure}
                \hfill
                \begin{subfigure}[c]{0.475\textwidth}  
                    \centering
                    \includegraphics[clip,width=\textwidth]{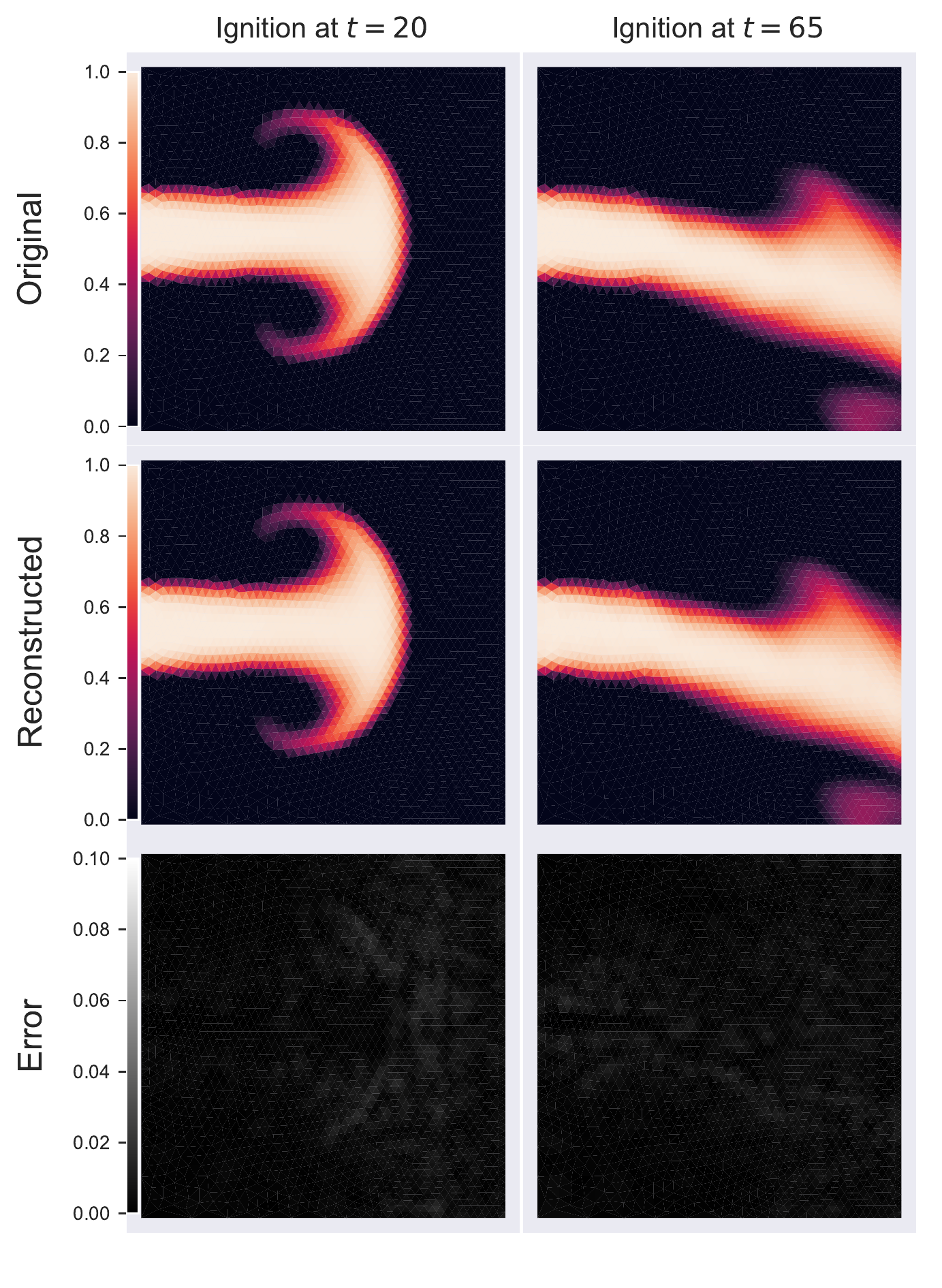}
                    \caption{\small QCNN compression.}
                    \label{fig:non-uniform-ignition-reconstruction}
                \end{subfigure}
                \caption{QCNN non-uniform ignition data compression.}
                \label{fig:qcnn_unstruct_ig}
            \end{figure}

            To maintain a benchmark of performance we interpolated the ignition data with 2D splines and evaluated the interpolant on the non-uniform mesh shown in \cref{fig:ignition_mesh}. This mesh contains 2189 points, more concentrated in the middle and the right-side of the domain, and was generated with \texttt{dmsh} \cite{dmsh_repo}. In order to evaluate the error introduced during the interpolation and subsequent re-sampling on the non-uniform mesh, we re-interpolated the data from the non-uniform mesh and compared it to the original data from the uniform grid. Using the same relative Frobenius-error averaged over all data samples, we found the interpolation process introduced $0.122\%$ error into the data.
            
            For this experiment, we continue to use a latent state \(\bm{z}\in\reals^{50}\), but our model varies slightly from \cref{sec:uniform-ignition} in the first and last QuadConv layers, although \cref{fig:pool-encoder} still describes the general structure. Because the input data is on a non-uniform mesh, the first QuadConv layer learns the quadrature weights and re-samples to a uniform grid. Doing this also allows us to use the max pooling operation and its adjoint to up-sample and down-sample, respectively. It should be noted that after re-sampling the data to a uniform grid and before sampling back to a non-uniform mesh, one could employ standard 2D convolutions. This is certainly a valid approach, but in this work we are seeking to validate the QuadConv operator itself, and not necessarily trying to optimize the architecture. The rest of the model remains unchanged until the last QuadConv layer in the decoder, where the data is sampled from the uniform grid back to the original non-uniform mesh.

            \edit{The GCN and SplineCNN in this experiment must similarly adapt their pooling layer to be well-defined on the non-uniform data. We use the KNN pooling operator found in \cite{Fey_Fast_Graph_Representation_2019} to interpolate the data from the mesh (after a graph convolution operation) back to a grid. This allows us to use the max pooling operator as defined on a grid just as in the QCNN. We use the KNN unpooling operator in \cite{Fey_Fast_Graph_Representation_2019} to then move back to the mesh in the second to last layer in the network. While there are other graph pooling operations, this is what we believe to be the most direct comparison possible with readily available software.}

            We will also compare against two other methods for dealing with the non-uniformity of the mesh data. First, we consider a voxel model, which is identical to the standard convolutional model used in \cref{sec:uniform-ignition} except that the input data is voxelized. That is to say, before the input data is passed through the network, it is aggregated onto a uniform grid. The inverse of this process, de-voxelization, is then applied after processing and before computing evaluation metrics. The second model we consider is a point-voxel model. This pairs the voxel model with a point-based branch where shared MLPs are applied directly to the point features themselves. See \cite{pointnet} and \cite{spvconv} for more details.
            \begin{table}[h]
                \centering
                \begin{adjustbox}{max width=\textwidth}
                    \begin{tabular}{lccccc}
                        \toprule
    \textbf{Model Type} & \textbf{Average Error} & \textbf{Max Error} & \textbf{Training Time (h)} & \textbf{\# of Steps} &\textbf{\# of Trainable Parameters} \\ 
    \midrule
    \textbf{Voxel} & 2.789\% & \hphantom{1}5.18\% & \hphantom{1}0.66 & 120,000 & 1,015,222 \\
    \textbf{Point-Voxel} & 0.956\% & 13.41\% & \hphantom{1}1.17 & 120,000 & 1,021,163 \\
    \textbf{GraphConv} & 0.730\% & 21.14\% & \hphantom{1}2.98 & 450,000 & 1,016,206 \\
    \textbf{SplineCNN} & 0.597\% & \hphantom{1}8.54\% & 19.34 & 3,600,000* & 1,025,710 \\
    \textbf{QuadConv} & 0.596\% & \hphantom{1}\textbf{2.05\%} & \hphantom{1}3.50 & 120,000 & 1,037,499 \\
                        \bottomrule
                    \end{tabular}
                \end{adjustbox}
                \caption{\small Results for non-uniform ignition data compression.}
                \label{tab:non-uniform-ignition-results}
            \end{table}
            As we can see in \cref{tab:non-uniform-ignition-results} and \cref{fig:non-uniform-ignition-reconstruction}, the QCNN produces similar accuracy as on the uniform grid. This suggests that our approximation is performing well over the non-uniform data found in the center of the domain, since we do not see an increase in error due to the change in the underlying structure of the data. Error histograms are similar to those in the uniform grid case and can be found in \cref{sec:appendix}. While the voxel model trains quite quickly, we can see that it is by far the worst performing method. This should not be altogether surprising, as the de-voxelization process introduces a hard lower bound on the reconstruction accuracy because mesh points within the same grid cell will necessarily have the same features. On the other hand, the point-voxel model is able to achieve comparable \edit{average error} to QuadConv, \edit{but its maximum error is quite high to due its poor generalization to the test data}. This speaks to QuadConv's ability to match \edit{and even exceed} other state-of-the-art methods for operating on non-uniform data. Importantly, we note that the shared MLPs of the point branch place a limit on how small a point-voxel network can be. Thus, such a method may not be suitable for data compression in a general setting. \edit{Finally, the GCN exhibits the same high maximum error as in the previous test case on the grid while performing well on average. While the SplineCNN is the best of all previous methods, its maximum error is still $6\%$ higher than QuadConv.}


        \subsection{Non-Uniform Mesh Flow Data}\label{sec:non-uniform-flow}
        
\begin{figure}[h]
                \centering
                \includegraphics[width=\textwidth]{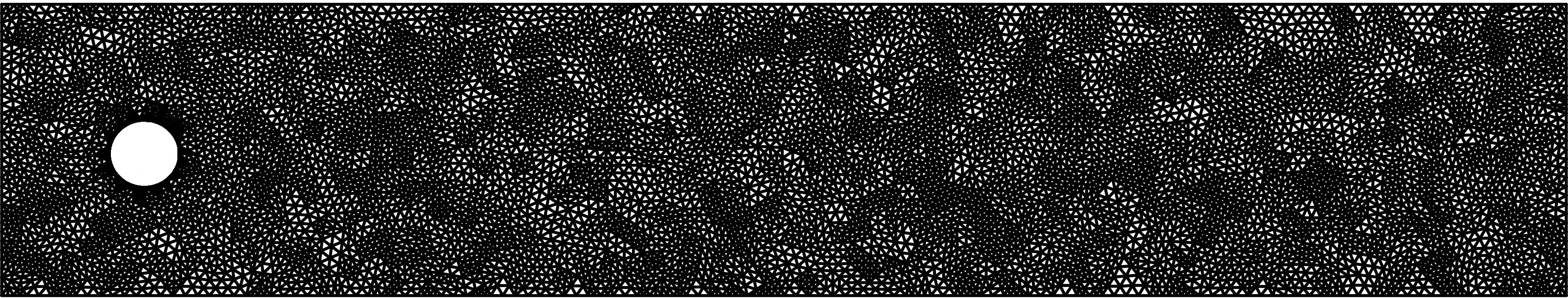}
                \caption{\small Non-uniform flow mesh.}
                \label{fig:flow-mesh}
            \end{figure}

            While the ignition examples are informative for verifying the performance of the QuadConv layer in comparison to standard convolution, the non-uniform mesh is not native to the data. We consider here an example of fluid flow around a cylinder at Reynolds number \(Re=100\) in a channel. This PDE simulation was conducted on the non-uniform mesh shown in \cref{fig:flow-mesh}. This dataset has 300 time points and the streamwise velocity at 7613 nodes in the domain. 
            
\begin{figure}[h]
                \centering
                \includegraphics[width=\textwidth]{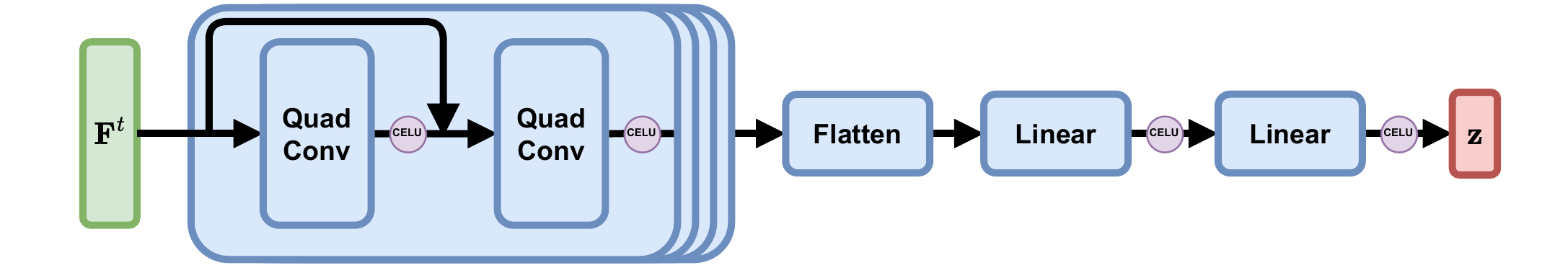}
                \caption{\small Encoder with four QuadConv based blocks. The linear layers comprise the MLP. The decoder is similarly structured, but mirrored to up-sample.}
                \label{fig:skip-encoder}
            \end{figure}

            This data is not as transport dominated as the jet ignition data, and, as a result, it is less challenging to compress. Our model is also substantially changed from the previous experiments. While we have thus far performed a $~50\times$ reduction in the data, this example has a latent state $\bm{z}\in\mathbb{R}^{15}$, which equates to a $~500\times $ reduction. This increase in data reduction is due to a different linear layer preceding the latent space, but it still requires effective feature extraction inside the QuadConv layers. Because of the non-Cartesian nature of the data, we no longer use the approach of sampling to a grid and applying max pooling layers. Instead, we use the QuadConv layers to down-sample, where a specified a number of the output points are sampled from the input points by sampling uniformly at random. \edit{The encoder structure is described in \cref{fig:skip-encoder}}. The random sampling technique is feasible in this scenario because the point density is fairly constant, but mostly we use it as a tool to show that other agglomeration procedures can be used for selecting non-uniform output points.
            \begin{table}[h]
                \centering
                \begin{adjustbox}{max width=\textwidth}
                    \begin{tabular}{lccccc}
                        \toprule
                        \textbf{Model Type} & \textbf{Average Error} & \textbf{Max Error} & \textbf{Training Time (h)} & \textbf{\# of Steps} &\textbf{\# of Trainable Parameters} \\ 
                        \midrule
                        \textbf{QuadConv} & 0.762\% & 2.24\% & 7.76 & 80,679 & 807,304 \\
                        \bottomrule
                    \end{tabular}
                \end{adjustbox}
                \caption{\small Results for non-uniform flow data compression.}
                \label{tab:non-uniform-flow-results}
            \end{table}

\begin{figure}[H]
                \centering
                \includegraphics[width=\textwidth]{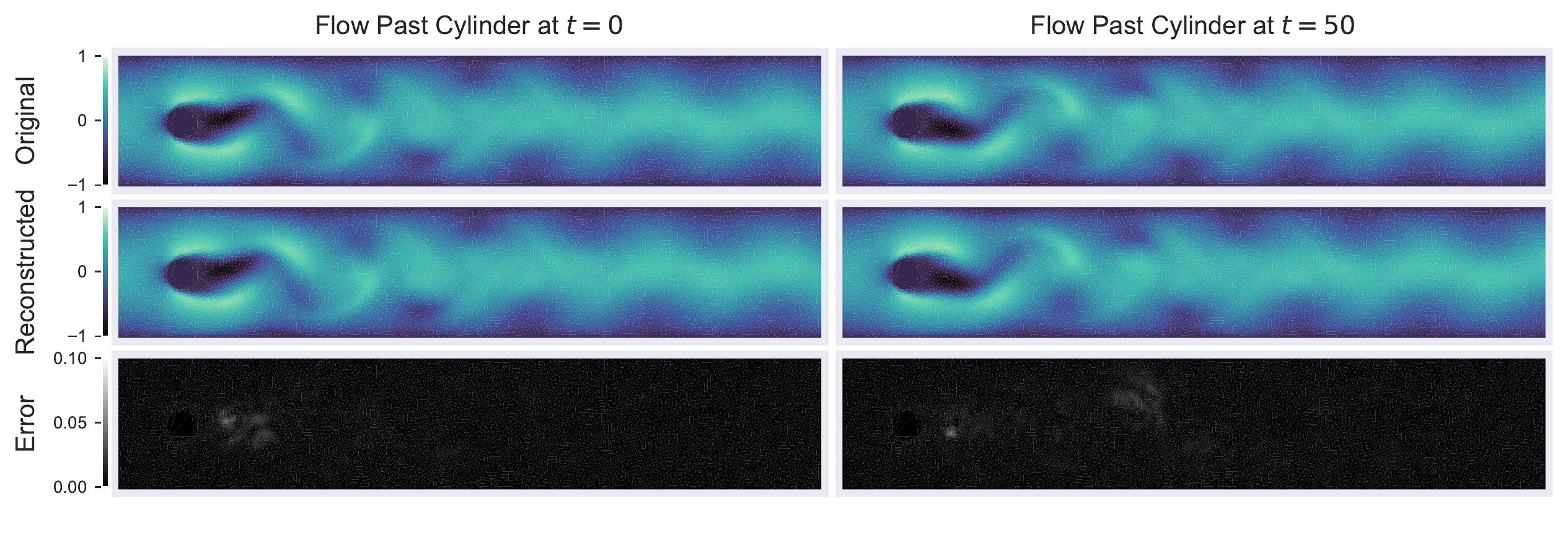}
                \caption{\small QCNN non-uniform flow compression.}
                \label{fig:non-uniform-flow-reconstruction}
            \end{figure}

            As shown in \cref{tab:non-uniform-flow-results} and \cref{fig:non-uniform-flow-reconstruction}, the flow data compresses well, with low error and converges considerably faster than the ignition data. This example shows that practical meshes are easily used, and high accuracy can be achieved with a QuadConv based network. The error histograms can be found in \cref{sec:appendix}.
            
            From \cref{fig:non-uniform-flow-reconstruction}, we observe very little degradation in the quality of the solution and no visible changes in the error. In fact, since the linear layers in the network dominate the total number of trainable parameters, our embedding of the data into a smaller space reduces the overall number of parameters in the network and slightly reduces training time. 

    
    \section{Discussion}\label{sec:discussion}
        In this work we have presented a new convolution operator for deep learning applications involving non-uniform data. Our operator approximates a continuous convolution via quadrature, and employs a learned continuous kernel to allow for arbitrary discretizations of the input data. In addition, we discussed an implementation of this operator that yields sufficiently reasonable time complexity for use in modern deep learning settings. Our experiments on both uniform and non-uniform PDE simulation data compression showed that, in practice, this quadrature convolution can match the performance of standard convolutions on uniform grid based data, and performs equally as well on non-uniform mesh data. 
    
    
        \subsection{Future Work}\label{sec:future-work}

            The results we have presented here demonstrate the promising performance of the QuadConv layer and QCNN in application, but there are still a number of interesting questions to address as to its performance and sensitivity to hyper-parameters. As discussed in \cref{sec:computation}, there are a number of ways to parameterize the map from point to filter, and we have only considered one such method here. A more thorough investigation into the alternative approaches is warranted. Also discussed in \cref{sec:methods}, and put to use in \cref{sec:experiments}, is the learning of the quadrature weights for a given set of input points. The accuracy of this learned quadrature itself can be measured separately using known data and kernel function. Understanding this may inspire regularization of the quadrature weights that could enable learning higher accuracy approximations. Mentioned briefly in \cref{sec:computation} was the idea of constructing the output points of a QuadConv layer via an agglomeration of the input mesh. This would allow one to maintain properties of the original mesh while still down-sampling. Such a method is more extensible than the random down-sampling used in \cref{sec:non-uniform-flow}, and would likely lead to better performance for QuadConv on meshes with more complex non-Cartesian geometries. Moreover, our continuous kernel allows us to learn representations of data on a continuous space which enables one to perform multi-level training. This would involve training on different resolutions of data using the same network but with variable quadrature weights in order to continue to approximate the convolution operator well while using the same kernels. Last but not least, developing online, parallel learning schemes to enable pass-efficient, ideally \textit{in situ}, data compression, as in \cite{dunton2020pass,pacella2022task}, is an important future research direction. 
        

    \section*{Acknowledgments}

        This material is based upon work supported by the Department of Energy, National Nuclear Security Administration under Award Number DE-NA0003968 as well as Department of Energy Advanced Scientific Computing Research Awards DE-SC0022283 and DE-SC0023346. AD's work has also been partially supported by AFOSR grant FA9550-20-1-0138. We would also like to thank Kenneth Jansen, John Evans, and Jeff Hadley from the University of Colorado Boulder for their helpful discussions surrounding this work.


    \bibliographystyle{elsevier/elsarticle-num}
    \bibliography{ref}
 
	
    \appendix
    \section{}\label{sec:appendix}

        \begin{figure}[H]
            \centering 
            \includegraphics[width=\textwidth]{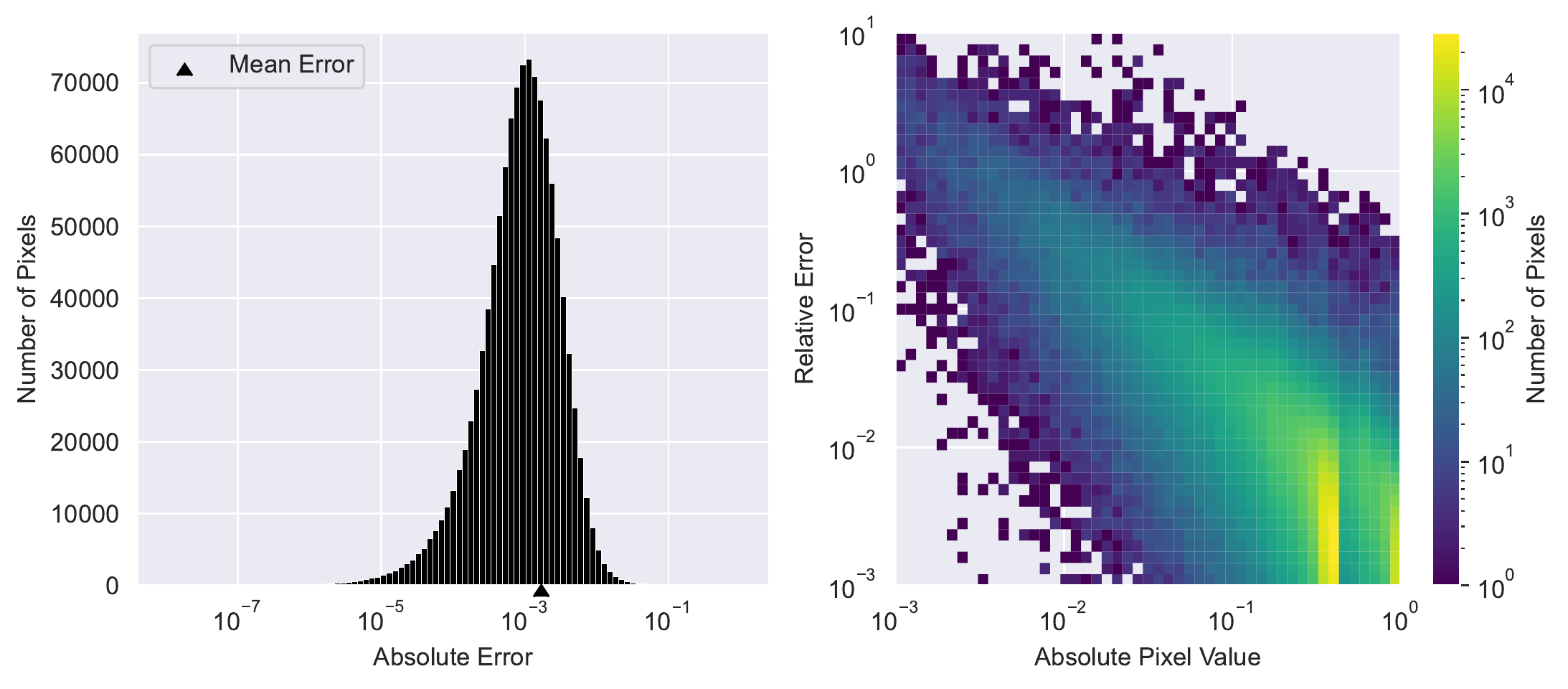}
            \caption{\small Error analysis of GCN for the uniform grid ignition data.}
            \label{fig:gcn-error}
        \end{figure}

        \begin{figure}[H]
            \centering 
            \includegraphics[width=\textwidth]{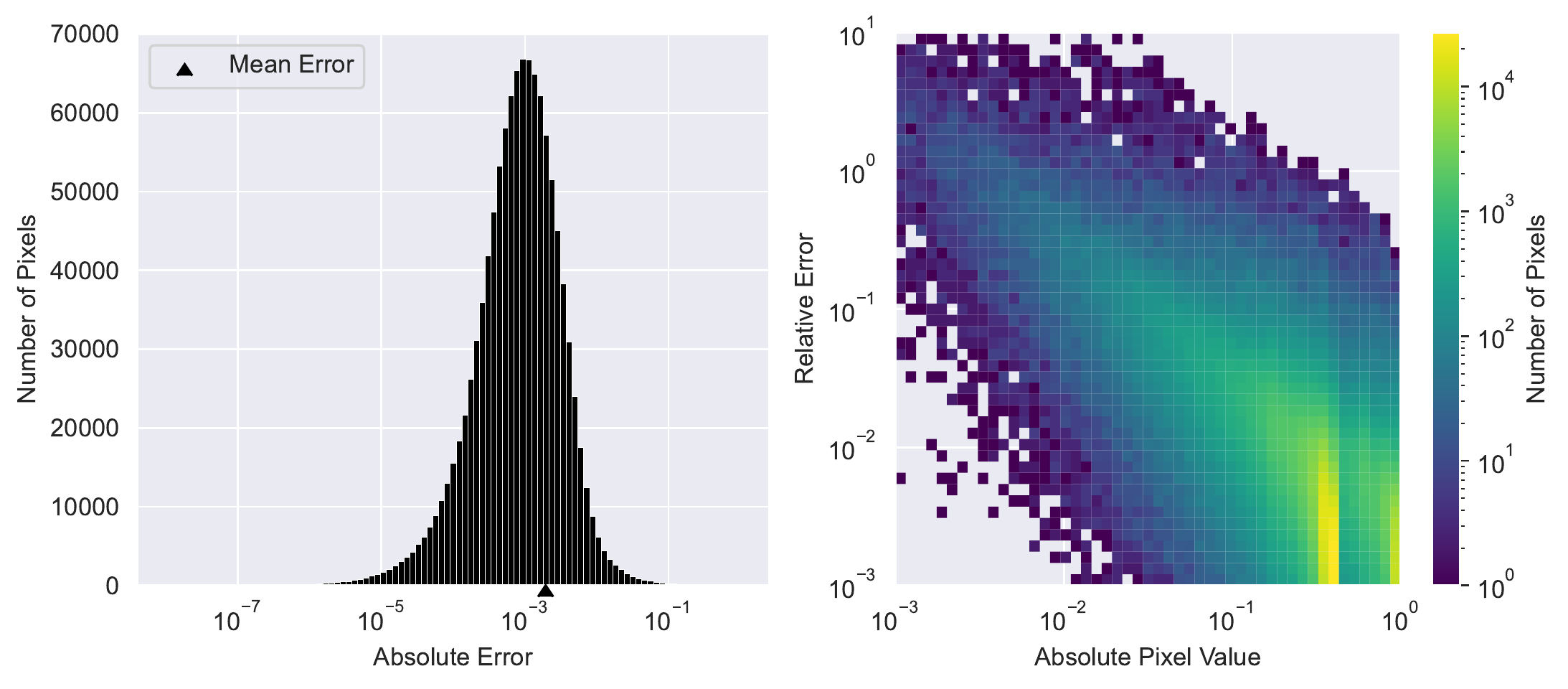}
            \caption{\small Error analysis of SplineCNN for the uniform grid ignition data.}
            \label{fig:splinecnn-grid-error}
        \end{figure}

        \begin{figure}[H]
            \centering
            \includegraphics[width=\textwidth]{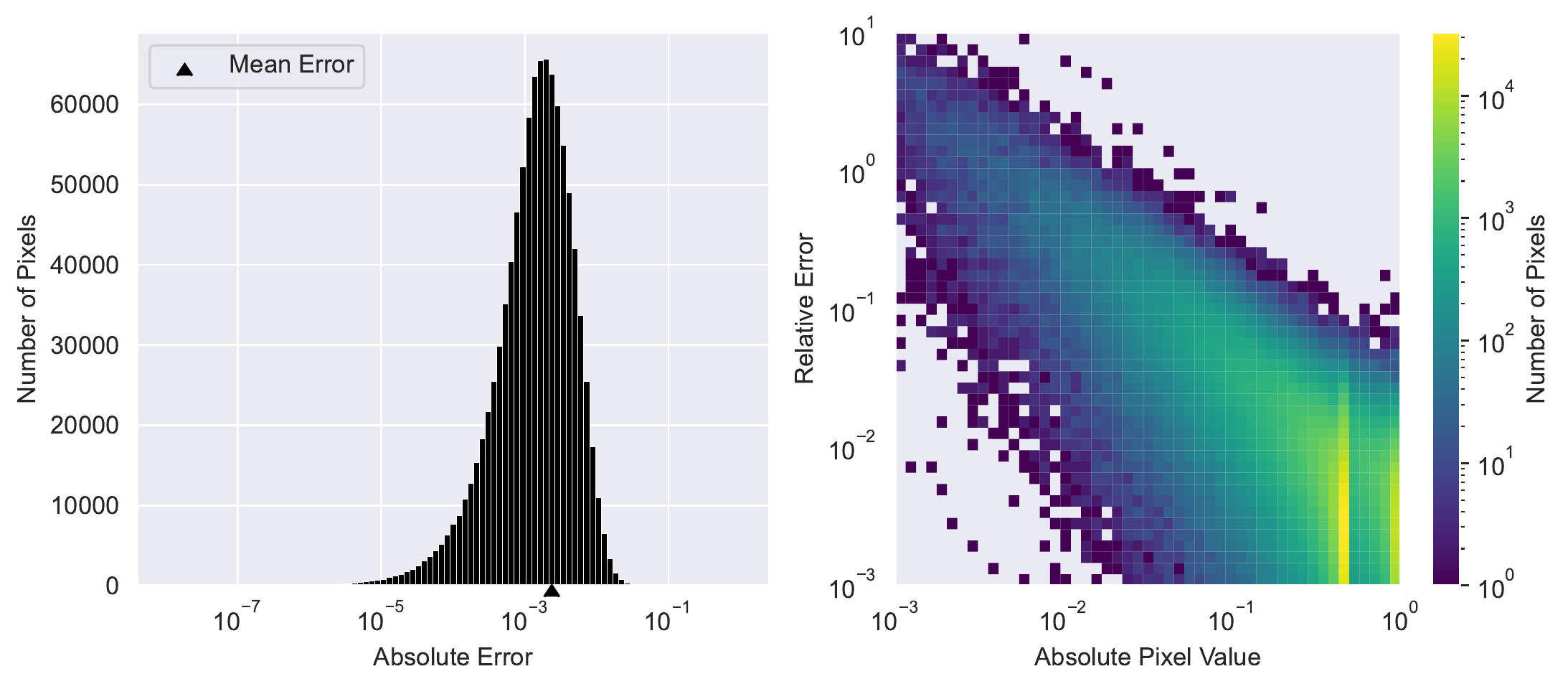}
            \caption{\small QCNN error analysis of non-uniform ignition data.}
            \label{fig:non-uniform-ignition-error}
        \end{figure}
    
        \begin{figure}[H]
            \centering 
            \includegraphics[width=\textwidth]{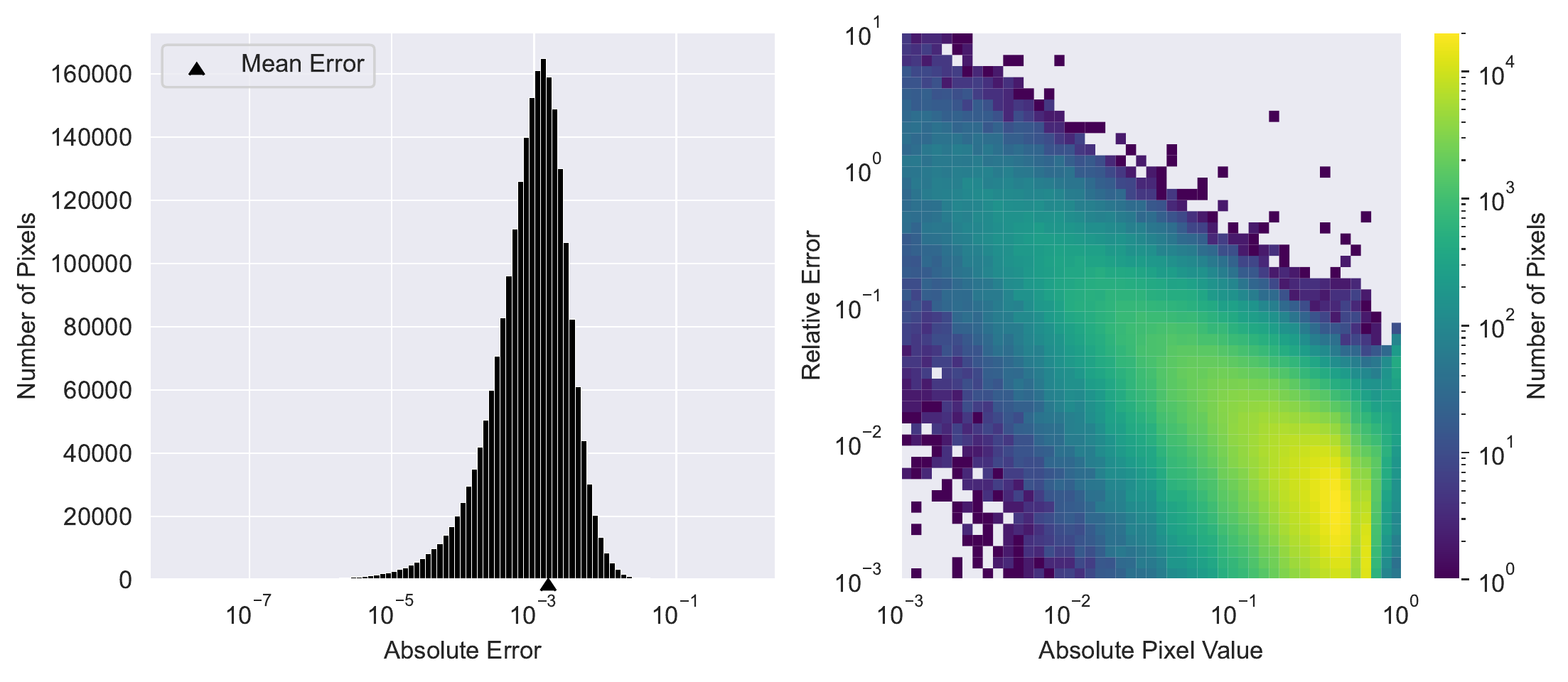}
            \caption{\small QCNN error analysis of non-uniform flow data.}
            \label{fig:non-uniform-flow-500-error}
        \end{figure}
    
\end{document}